%% file: main.tex
\documentclass[sigconf, nonacm]{acmart}

\usepackage[vlined,ruled,linesnumbered]{algorithm2e}
\usepackage{subcaption}
\usepackage{csquotes}
\usepackage{multirow}
\usepackage{url}
\usepackage{tabularx}

\usepackage{enumitem}
\usepackage{color,soul}
\usepackage{amsmath,amsfonts,amsthm,color,mathtools}
\newtheorem{definition}{Definition}
\newtheorem{exam}{Example}
\usepackage{hyperref}
\usepackage{wrapfig}
\usepackage{graphicx}
\usepackage{booktabs}

\newcommand{\V}{\mathcal{V}}
\newcommand{\E}{\mathcal{E}}
\newcommand{\eat}[1]{}
\newcommand{\spara}[1]{\smallskip\noindent{\bf #1}}

\usepackage{url,hyperref,lineno,microtype}

\begin{document}
\title{Data Depth and Core-based Trend Detection on Blockchain Transaction Networks}

%%
%% The "author" command and its associated commands are used to define the authors and their affiliations.
\author{Jason Zhu}
\affiliation{%
  \institution{University of Manitoba}
  \city{Winnipeg}
  \state{Canada}
}
\email{zhuj3410@myumanitoba.ca}

\author{Arijit Khan}
\affiliation{%
  \institution{Aalborg University}
  \city{Aalborg}
  \state{Denmark}
}
\email{arijitk@cs.aau.dk}

\author{Cuneyt Gurcan Akcora}
\affiliation{%
  \institution{University of Central Florida}
  \city{Orlando}
  \state{USA}
}
\email{cuneyt.akcora@ucf.edu}

\begin{abstract}
\input{sections/00_abstract.tex}
\end{abstract}

\maketitle

\input{sections/01_introduction.tex}
\input{sections/02_related_work.tex}
\input{sections/03_preliminaries_rw.tex}

\input{sections/04_data_depth.tex}
\input{sections/05_methodology.tex}

\input{sections/06_experiments.tex}
\input{sections/07_conclusion.tex}

%\clearpage

\bibliographystyle{ACM-Reference-Format}
\bibliography{main}

\end{document}

%% file: sections/00_abstract.tex
Blockchains are significantly easing \textit{trade finance}, with billions of dollars worth of assets being transacted daily. However, analyzing these networks remains challenging due to the sheer volume and complexity of the data.
We introduce a method named {\textsf InnerCore} that detects market manipulators within blockchain-based networks and offers a sentiment indicator for these networks. This is achieved through data depth-based core decomposition and centered motif discovery, ensuring scalability. 
{\textsf InnerCore} is a computationally efficient, unsupervised approach suitable for analyzing large temporal graphs. 
We demonstrate its effectiveness by analyzing and detecting three recent real-world incidents from our datasets: the catastrophic collapse of LunaTerra, the Proof-of-Stake switch of Ethereum, and the temporary peg loss of USDC -- while also verifying our results against external ground truth. Our experiments show that {\textsf InnerCore} can match the qualified analysis accurately without human involvement, automating blockchain analysis in a scalable manner, while being more effective and efficient than baselines and state-of-the-art attributed change detection approach in dynamic graphs.

%% file: sections/01_introduction.tex
\section{Introduction}
\label{sec:intro}
Blockchain technology \citep{nakamoto2008bitcoin,wood2014ethereum} is revolutionizing the way we store and transfer digital assets {in multiple domains including internet-of-things \citep{LiuZYZTZ23}, healthcare \citep{LiuYAXZT23}, and digital evidence \citep{TianLQSS19}}. Public blockchain networks are completely open, allowing anonymous addresses to utilize transactions for cryptocurrency movement and asset trading/investment. While the technology offers numerous benefits, it poses significant challenges, particularly in the area of cybersecurity. Blockchains enable electronic crimes in a variety of ways \citep{abs-2212-13452}, ranging from demands for ransomware \citep{huang2018tracking} to transactions in darknet markets \citep{jiang2021illicit}.

One of the biggest challenges in securing blockchain networks is detecting and preventing e-crime. E-crime detection requires scalable analysis of large-scale blockchain graphs in real-time, where results are both qualified and manageable by human analysts. To address this challenge, researchers have developed tools and algorithms for analyzing blockchain networks \citep{victor2021alphacore,akcora2017chainlet,SuG022,KA22}.

Unfortunately, analyzing blockchain networks is an arduous task, given their large size and the involvement of anonymous actors. It is crucial to devise scalable and effective methods that can analyze blockchain networks in real-time, to preempt future losses. The failure to conduct a timely analysis of blockchain networks has already resulted in a staggering loss of billions of dollars to blockchain users, as exemplified by the recent downfall of LunaTerra \citep{lunaterra}.

In this article, we introduce a new approach to detecting e-crimes and trends detection. Our approach, {\textsf InnerCore}, involves identifying influential addresses with data depth-based core decomposition and further filtering out the role of addresses by using centered motifs. 
{\textsf InnerCore} analysis reduces large graphs having more than 400K nodes and 1M edges to an induced subgraph of less than 300 nodes and 90K edges, while still being able to detect the influential nodes.
{\textsf InnerCore} is unsupervised and highly scalable, yielding only $\sim$4-second  running times on daily Ethereum graphs with $\sim$500K nodes and $>$1M edges. We apply {\textsf InnerCore} to three recent important events in the blockchain world: the collapse of LunaTerra in May 2022, the Proof-of-Stake (PoS) switch of Ethereum in September 2022,
and the temporary peg loss of USDC in March 2023. Experimental results demonstrate that our proposed approach effectively detects significant changes in the network without human intervention. Moreover, {\textsf InnerCore} excels in accurately identifying market-manipulating addresses within the network, underscoring its effectiveness in pinpointing key actors.

Our key novelties and contributions are summarized below.
\begin{itemize}[leftmargin=.25in]
 \item 
  \textit{InnerCore}: We propose {\textsf InnerCore}, a data depth-based core discovery method that can identify the influential 
 traders in blockchain-based asset networks (\S\ref{sec:methinnercore}).
 \item 
  \textit{Explainable behavior}: We develop two metrics, {\textsf InnerCore} expansion and decay (\S\ref{sec:methdecayexpansion}), that provide a sentiment indicator for the networks and explain trader mood (\S\ref{sec:methpatterns}). 
 \item 
  \textit{Unsupervised address discovery}: Through conducting node ranking with a centered-motif approach in temporal asset networks, we demonstrate that {\textsf InnerCore} tracking detects market manipulators and e-crime behavior and warns the network about possible long-term instability, without the need for supervised address discovery (\S\ref{sec:methmotif}).
 \item 
  \textit{Scalability}: Due to their computational efficiency and ability to utilize only a small portion of graph nodes and edges to analyze overall behavior, the {\textsf InnerCore} discovery and expansion/decay calculations are suitable on large temporal graphs including Ethereum transaction and stablecoin networks. {\textsf InnerCore} is more effective and efficient than baselines \citep{victor2021alphacore,BatageljZ11} and the state-of-the-art attributed change detection method in dynamic graphs \citep{huang2023fast}  (\S\ref{sec:exp}).
\end{itemize}
%
%We discuss preliminaries and our problem in \S\ref{sec:prelim}.
%Additional related works are specified in the Appendix. %our full version~\cite{InnerCoreAppendix}.

%% file: sections/02_related_work.tex
\section{Related Work}
In recent years, several studies focused on analyzing different aspects of the blockchain networks \citep{ChenWZCZ19,AkcoraLGK20,kalodner2017blocksci,GuidiM20}, particularly in the Ethereum network. 
Researchers working on natural
language processing and sentiment analysis using tweets, news articles, cryptocurrency prices, and charts, Google Trends about blockchains \citep{VNO19,KS20} could find supporting evidence based on
blockchain data analysis. 
\citet{oliveira2022analysis} performed an analysis of the effects of external events on the Ethereum  platform, highlighting short-term changes in the behavior of accounts and transactions on the network. \citet{aspembitova2021behavioral} used temporal complex network analysis to determine the properties of users in the Bitcoin and Ethereum markets and developed a methodology to derive behavioral types of users.

Other studies focused on specific aspects of the Ethereum network. For instance, \citet{casale2021networks} analyzed the networks of Ethereum Non-Fungible Tokens using a graph-based approach, while \citet{silva2020characterizing} characterized relationships between primary miners in Ethereum using on-chain transactions. Meanwhile, \citet{victor2019measuring} measured Ethereum-based ERC20 token networks, and \citet{kiffer2018analyzing} examined how contracts in Ethereum are created and how users interact with them.

{Numerous researchers found success in anomaly detection through the strategic exploration of the Ethereum transaction network using graph representation. In particular, \citet{patel2020springer} proposed an one-class graph neural network-based anomaly detection framework for Ethereum transaction networks that harnesses graph representation. \citet{wu2023ieeetran} proposed a scalable transaction tracing tool which incorporates a biased search method to guide the search of fund transfer traces on transaction graphs.}

\citet{zhao2021temporal} investigated the evolutionary nature of Ethereum interaction networks from a temporal graph perspective, detecting anomalies based on temporal changes in global network properties and forecasting the survival of network communities using relevant graph features and machine learning models. \citet{li2021measuring} analyzed the magnitude of illicit activities in the Ethereum ecosystem using proprietary labeling data and machine learning techniques to identify additional malicious addresses. \citet{kilicc2022fraud} predicted whether given addresses are blacklisted or not in the Ethereum network using a transaction graph and local and global features. 

Our temporal approach for analyzing the effects of external events on a blockchain platform is similar to the one used by \citet{anoaica2018quantitative}. The authors examined the temporal variation of transaction features in the Ethereum network and observed an increase in activity following the announcement of the Ethereum Alliance creation.  \citet{gaviao2020transaction} also studied the evolution of users and transactions over time, showing the centralization tendency of the transaction network. \citet{kapengut2022event} studied the Ethereum blockchain around the BeaconChain phase of the PoS transition (September 15, 2022), but the authors focused on the power efficiency and miners' rewards around the transition.

Finally, \citet{khan2022graph} conducted a survey of datasets, methods, and future work related to graph analysis of the Ethereum blockchain data, while Poursafaei's PhD thesis \citep{ramezan2022anomaly} presented results on temporal anomaly detection in blockchain networks.

%% file: sections/03_preliminaries_rw.tex
\section{Background and Problem}
\label{sec:prelim}
We discuss preliminaries on blockchain and stablecoins (\S \ref{sec:block}, \S \ref{sec:stable}), followed by one key technique {\textsf AlphaCore} decomposition based on data depth (\S \ref{sec:alpha}). We introduce our problem in \S \ref{sec:prob}.
%
%\vspace{-2mm}
\subsection{Blockchain and Smart Contracts}
\label{sec:block}
A blockchain is an immutable public ledger that records transactions in discrete data structures called blocks. The earliest blockchains are cryptocurrencies such as Bitcoin and Litecoin where a transaction is a transfer of coins.
The Ethereum project \citep{wood2014ethereum} was created in July 2015 to provide smart contract functionality on a blockchain. Smart contracts are Turing complete software codes, replicated across a blockchain network, ensuring deterministic code execution and can be verified publicly. 
Smart contracts have implemented mechanisms to trade digital assets, known as tokens \citep{victor2019measuring}. Similar to cryptocurrencies, a token is transferred publicly between accounts (addresses), and may have an associated value in fiat currency which is arbitrated by token demand and supply in the real world.

\smallskip 

{\noindent\textbf{Blockchain Transaction Network vs. Mining Network.} In blockchain transaction networks, the nodes represent individual participating addresses within the network, while the edges signify the actual transactions involving transfer of assets between these addresses. On the other hand, in blockchain mining networks, nodes are computational entities that play a crucial role in maintaining blockchain integrity by validating and appending transactions to the ledger through a consensus mechanism. We focus on blockchain transaction networks, where edges are directed and weighted. An edge weight corresponds to the numerical value associated with the edge incident to a node. For instance, in a blockchain token transcation network, the numerical value denotes the amount of token sent from one address
to another.}

%
%\vspace{-2mm}

\subsection{Stablecoins}
\label{sec:stable}
A stablecoin is a smart contract-based asset whose price is protected against volatility by i) collateralizing the stablecoin with one or more offline real-life assets (e.g., USD, gold), ii) using a dual coin, or by iii) employing algorithmic trading mechanisms \citep{moin2019classification,LI2024103747}. 

In the \textit{pegged asset mechanism}, an increase in the price is countered by creating more stablecoins (i.e., coin minting) and selling them to traders at the pegged price. The \textit{dual coin mechanism} operates by having a management coin, referred to as the dual coin, to oversee a stablecoin. The traders of the dual coin participate in decision making through voting and receive benefits from the stablecoin's transactions. In the event that the stablecoin's price rises, some of the dual coin will be sold to purchase and decrease the supply of the stablecoin. Conflicting demand and supply dynamics of the two coins are assumed to stabilize the stablecoin's price.  However, traders may lose faith in the stablecoin to such a degree that they might also not buy the dual coin, however cheap it becomes.
Stablecoins that are based on \textit{algorithmic} trading do not require collateral for stability. They achieve stability through the utilization of a blockchain-based algorithm that adjusts the supply of tokens automatically in response to changes in demand.

{It is worth noting that for an Ethereum token such as the UST {(TerraUSD)} stablecoin, there can be at most $k$ tokens issued within this network, with the value of $k$ being set by the project owner, subject to the condition that it must be $\leq (2^{256}-1)$ (due to Ethereum virtual machine operating on 256 bit words). Furthermore, each of these $k$ tokens can be subdivided into a maximum of $10^{18}$ subunits (an Ethereum protocol specified value). Therefore, the total subunit capacity for a token within the system is $k\times 10^{18}$ subunits.}

 %
%\vspace{-2mm}
\subsection{Data Depth-based Core Decomposition}
\label{sec:alpha}
Core decomposition \citep{Malliaros20} is a central technique used in network science to determine the significance of nodes and to find community structures in a wide range of applications such as biology~\citep{luo2009core}, social networks~\citep{al2017identification}, and visualization~\citep{zhang2012extracting}.
One of the best-known representatives of core decomposition algorithms, graph-$k$-core~\citep{seidman1983network,BatageljZ11}, finds the maximal subgraph where each node has at least $k$ neighbors in that subgraph.
Although the graph-$k$-core algorithm demonstrates high utility for the analysis of graph structural properties, it does not account for important graph information such as the direction of edges, edge weights, and node features.

To address these limitations, modifications to graph-$k$-core have been proposed, %to tackle task-specific graphs, 
e.g., graph-$k$-core in weighted and directed graphs, generalized $k$-core \citep{al2017identification,Zhou0HY0021,LiaoLJHXC22,batagelj2002Generalized,Garas_2012,GiatsidisTV11}. Different from them, \\ {\textsf AlphaCore}~\citep{victor2021alphacore} is a recent core decomposition algorithm that combines multiple node properties using the statistical methodology of data depth ~\citep{mosler2012multivariate}. The key idea of data depth is to offer a center-outward ordering of all observations by assigning a numeric score in $(0,1]$ to each data point with respect to its position within a cloud of a multivariate probability distribution. Using such a data depth function designed for directed and weighted graphs, {\textsf AlphaCore} maps a node with multiple features to a single numeric score, while preserving its relative importance with respect to other nodes.   

%As the {\textsf AlphaCore} decomposition unfolds, the data depth values are repeatedly updated through the calculation of node property functions and the application of data depth to the resulting values.

Consider a directed and weighted multigraph, $G(\V,\E,w)$, where $\V$ represents the set of nodes and $\E$ is a multiset of edges. The weight of each edge is designated by the weight function $w : \E \rightarrow \mathbb{R}^+$. In accordance with the generalized core definitions introduced in~\citet{batagelj2002Generalized}, a node property function can assign a real value to each node $v \in \V$, based on edge properties such as weight and node features.  A node $v$ can be represented by its feature vector $\textbf{x}\in \mathbb{R}^d$, where $d$ features have been computed for the node $v$.

\begin{definition}[Mahalanobis depth to the origin (MhDO)]
Let $\textbf{x}\in \mathbb{R}^d$ be an observed data point, then Mahalanobis (MhD) depth of $\textbf{x}$ in respect to a $d$-variate probability distribution $F$ with mean vector $\mu_F \in \mathbb{R}^d$ and covariance matrix $\Sigma_F \in \mathbb{R}^{d\times d}$ is given by
 \begin{equation}
MhDO_F(\textbf{x})=\bigl(1+\textbf{x}^\top\Sigma^{-1}_F\textbf{x}\bigr)^{-1},
 \end{equation}
 $\Sigma_F$ is the covariance matrix of $F$. The Mahalanobis data depth to origin (MhDO) measures the degree of ``outlyingness" of point $\textbf{x}$ (in this context, the node property column vector) in relation to origin $\mathbf{0}$.
 \label{def:mhdo}
 \end{definition}
As the {\textsf AlphaCore} decomposition unfolds, the core value $\alpha$ of a node is established using a data depth threshold $ \epsilon\in [0,1]$ that is applied to remove neighboring high-depth nodes iteratively. Nodes with high property values, such as large edge weights, generally have a low depth, while nodes with low property values often have a high depth, such as most blockchain nodes that trade small amounts of tokens. However, node property values are not the only factor that determines depth; the community structure around the node also plays a role. Nodes are considered to be in the $\alpha = (1-\epsilon)$-core if their depth, relative to themselves, is no more than $\epsilon$.

\smallskip 

\noindent\textbf{Why Data Depth?} Data depth provides a more precise identification of crucial nodes compared to state-of-the-art core decomposition algorithms and acts as a combination of centrality measure and core decomposition~\citep{victor2021alphacore}. Unlike traditional decomposition algorithms, a depth-based decomposition does not require the specification of multiple feature weighting parameters to perform effectively on a particular task. %We provide a running example in the Appendix.  

\begin{figure}
    \centering \includegraphics[width=1\linewidth]{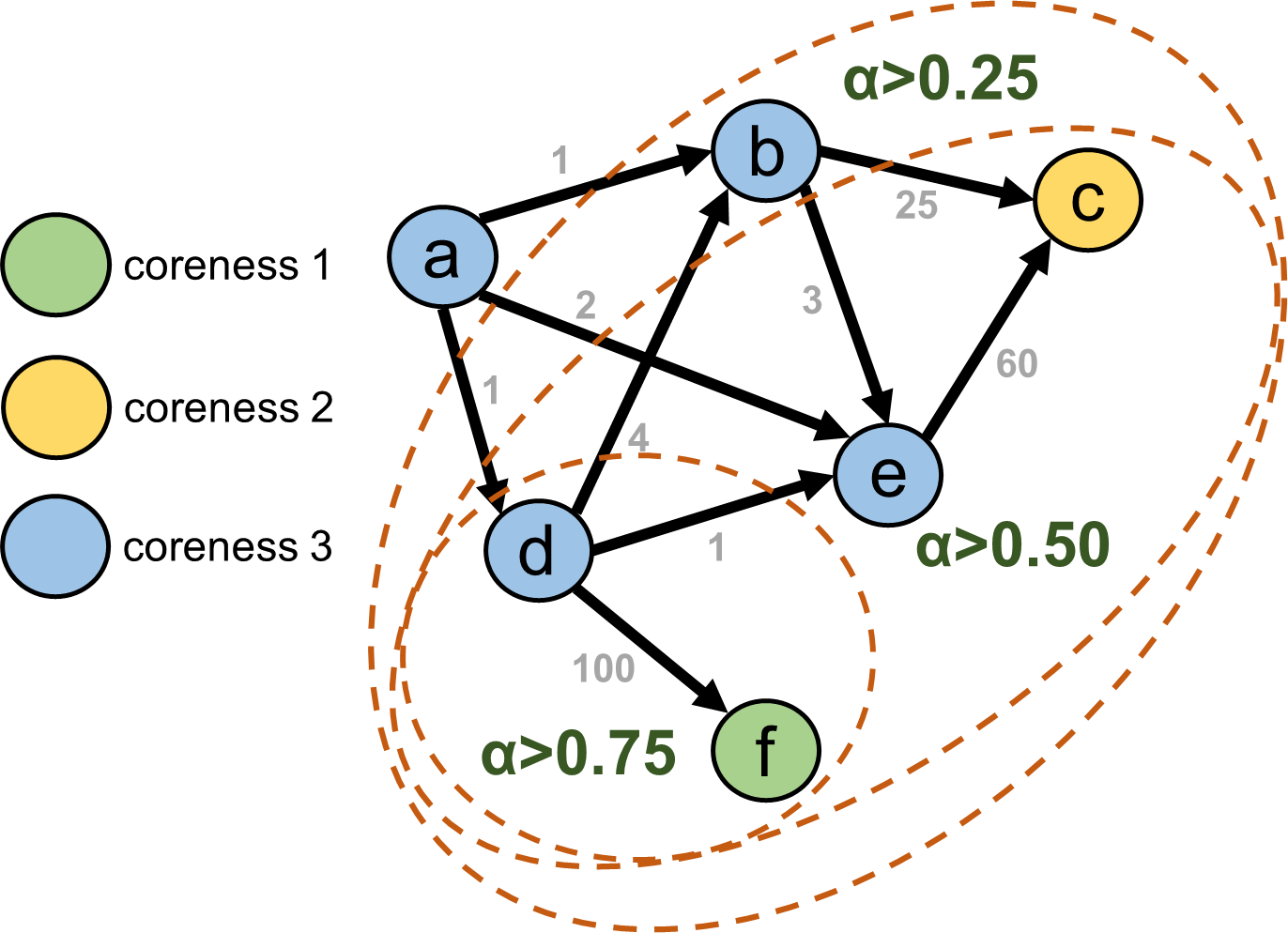}
    \vspace{-3mm}
    \caption{\small A running example to compare between the graph-$k$-core and {\textsf AlphaCore} decomposition methods. The Coreness of nodes according to graph-$k$-core decomposition is shown with different node colors, whereas {\textsf AlphaCore} is run with in-strength and out-strength as node features with a step size of 0.25. Different {\textsf AlphaCore}s are shown using dotted boundaries. \label{fig:running_example}}
    % \vspace{-5mm}
\end{figure}
%
%\vspace{-2mm}

\input{sections/033_aphacore_example}

%Our focus is also on detection rather than prediction, emphasizing the importance of identifying and addressing malicious activities within blockchains. 

%To resolve above problems, we identify nodes in the innermost core, as well as characterize three-node motifs in these innermost cores from our transaction networks.
%In our experiments, we demonstrate that nodes in the innermost core are more useful, compared to other notions of important nodes, in characterizing and predicting the future success of the blockchain assets. 

%% file: sections/033_aphacore_example.tex
\spara{An Example of AlphaCore.}
To better illustrate the differences between the traditional graph-$k$-core and {\textsf AlphaCore} decomposition methods, we showcase an example in Figure \ref{fig:running_example}.  %We are interested in the inner core that captures only the most important traders of a financial network.  
In the case of graph-$k$-core, the innermost core is the 3-core, whereas the {\textsf InnerCore} of {\textsf AlphaCore} would be the core of $\alpha$ $>$ 0.75.  Note that the 3-core consists of nodes that trade frequently with themselves, but their trade volumes with themselves are not that significant compared to other transactions which exist in the network.  In certain analyses of financial networks such as anomalous address detection, being able to filter out these negligible transactions and their participating nodes, while still capturing more meaningful ones, significantly improves the accuracy and scalability of subsequent computations on the decomposed network core.  On the other hand, the {\textsf AlphaCore} of $\alpha$ $>$ 0.75 is able to capture both the nodes that participate in the largest transactions which occur in the example network, while filtering the negligible transactions and their participating nodes.  
We point out that the main limitation with graph-$k$-core is that it only considers node degrees, whereas {\textsf AlphaCore} is flexible and can consider any combination of node features as outlined in Table 1, without requiring to specify any feature weighting parameters to perform effectively on a particular task.  
Therefore, in networks where edge weights fall under a broad range and they are meaningful distinguishing factors, we recommend {\textsf AlphaCore} over the traditional graph-$k$-core decomposition.

\subsection{Problem Definition}
\label{sec:prob}
Given a  weighted, directed, multi-graph {representation of a blockchain transaction network} over successive timestamps, where $G_t(\V_t,\E_t, w_t)$ denotes the graph at timestamp $t$, $\V_t$ its set of nodes (traders \footnote{{While a
trader can own multiple addresses, a typical trader has a main address that
holds the bulk of the assets. With our {\textsf InnerCore} approach, we are interested in capturing these main
addresses (i.e., nodes) and the respective traders.}}, exchanges, liquidity pools, etc.), and $\E_t$ multiset of edges (i.e., transactions) representing {the amount of asset transferred between two nodes},  {\bf (i)} detect the node set $S_t \subseteq \V_t$ at time $t$ such that the behavior of nodes in $S_t$ can %be used to 
characterize the future success of the underlying asset at $t^\prime>t$, and {\bf (ii)} categorize nodes' behavior in terms of the future health and success of the underlying asset. 
%{We define traders as externally owned addresses ({\textsf EoA}s).}

\noindent\textbf{E-crime Detection vs. Prediction.} In blockchain space, predictions can only go so far, as we are unable to anticipate malicious transactions that originate from the external world. At most, what we can do is to detect e-crime transactions among the vast number of transactions taking place. This detection process is highly valuable because when a significant crime occurs, we have access to public graphs of the affected assets. However, the sheer volume of addresses and transactions makes qualitative analysis impractical. This is where blockchain data analytics tools come into play, aiming to narrow down the search space by providing a ranking of maliciousness to addresses and transactions.

%% file: sections/04_data_depth.tex
\section{Data Depth}
\label{sec:depth}
Depth functions have been initially introduced in the setting of non-parametric multivariate analysis to define affine invariant versions of median, quantiles, and ranks in higher dimensional spaces where there is no natural order  (see historical overviews by~\citet{mosler2012multivariate,Nieto-Reyes:Battey:2015}).  The key idea of the depth approach is to offer a center-outward ordering of all observations by assigning a numeric score in $(0,1]$ range to each data point with respect to its position within a cloud of multivariate or functional observations or a probability distribution. Nowadays, data depth is a rapidly developing field that gains increasing momentum due to the wide applicability of depth concepts to classification, visualization, high dimensional and functional data analysis~\citep{Hyndman:Shang:2010, Narisetty:Nair:2016, mozharovskyi2020nonparametric, sguera2020notion, zhang2021depth}.
Most recently, depth approaches have found novel applications in density-based clustering and space-time data mining~\citep{Jeong:etal:2016, HuangGel2017, vinue2020robust}, shape recognition and uncertainty quantification in computer graphics~\citep{Whitaker:etal:2013, sheharyar2019visual}, ordinal data analysis~\citep{Kleindessner:vonLuxburg:2017} and computational geometry for privacy-preserving data analysis~\citep{mahdikhani2020achieve}.
Nevertheless, data depth is yet a largely unexplored concept in network sciences~\citep{Fraiman:et:2015, raj2017path,Tian:Gel:2017, tian2019fusing}.

\begin{definition}[Data Depth] Formally, let $E$ be a Banach space (e.g., $E=\mathbb{R}^d$), $\mathcal B$ its Borel sets in $E$, and $\mathcal P$ be a set of probability distributions on $\mathcal B$. We view $\mathcal P$ as the class of empirical distributions giving equal probabilities $1/n$ to $n$ data points in $E$. Then, a data depth function is a function $\mathbb{D}: E\times \mathcal P \longrightarrow [0,1]$, $(x,P) \longrightarrow \mathbb{D}(x|P)$, $x\in E, P\in \mathcal P$ that satisfies the following desirable properties: \textit{affine invariant}, \textit{upper semi-continuous} in $x$,
\textit{quasiconcave} in $x$ (i.e., having convex upper level sets) and \textit{vanishing as} $||x||\to \infty$. Specifically, a data depth function $\mathbb{D}(x)$ measures how closely an observed point $x \in \mathbb{R}^d$, $d\geq 1$, is located to the center of a finite set $\mathcal{X}\in \mathbb{R}^d$, or relative to $F$, which is a probability distribution in $\mathbb{R}^d$. In complex network analysis, these points may correspond to nodes or edges having features.
\end{definition}

Among many depth functions formulated to date, the Mahalanobis depth is one of the most prominent in the current practice.

\begin{definition}[Mahalanobis (MhD) depth]
Let $x\in \mathbb{R}^d$ be an observed data point, then Mahalanobis (MhD) depth of $x$ with respect to a $d$-variate probability distribution $F$ having mean vector $\mu_F \in \mathbb{R}^d$ and covariance matrix $\Sigma_F \in \mathbb{R}^{d\times d}$ is given by
 \begin{equation}
MhD_{\mu_F}(x)=\bigl(1+(x-\mu_F)^\top\Sigma^{-1}_F(x-\mu_F)\bigr)^{-1}.
 \end{equation}
Here $^\top$ denotes matrix transpose. The MhD depth measures the \textit{outlyingness} of the point with respect to the deepest point of the distribution (here $\mu_F$), and allows to easily handle the elliptical family of distributions, including a Gaussian case.
\end{definition}
MhD offers flexibility in changing the reference point with respect to which we compute data rankings. For instance, instead of $\mu_F$ we can select an arbitrary point $x_0\in \mathbb{R}^d$ and compute MhD in respect to this new reference point $x_0$
\begin{equation}
\label{MhD_arb}
MhD_{x_0}(x)=\bigl(1+(x-x_0)^\top\Sigma^{-1}_F(x-x_0)\bigr)^{-1}.
 \end{equation}
Furthermore, $\Sigma_F$ can be substituted by any empirical estimator of covariance matrix $\hat{\Sigma}$ obtained from the observed data sample $x_1, x_2, \ldots, x_n$.

%% file: sections/05_methodology.tex
\section{Methodology}
\label{sec:method}
\begin{figure}
    \centering \includegraphics[width=1\linewidth]{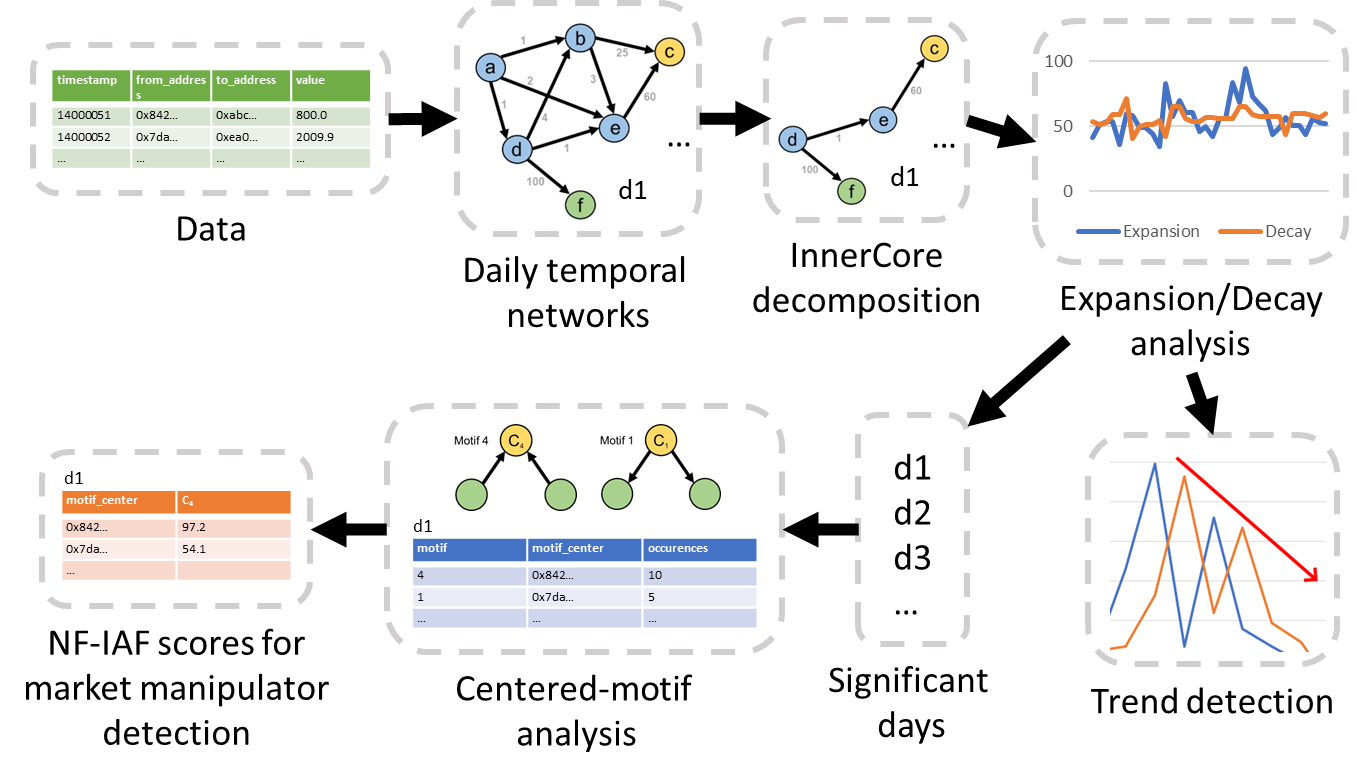}
    \vspace{-5mm}
    \caption{\small Flowchart of our methodology for identification of significant days and subsequent anomalous addresses.
    }
    \label{fig:methodology}
    % \vspace{-4mm}
\end{figure}

 Our methodology is illustrated in Figure~\ref{fig:methodology}. In keeping with the routine of daily life, blockchain {transaction} networks are frequently examined on a 24-hour  basis~\citep{casale2021networks,chen2020understanding}. We divide a blockchain {transaction} network into daily intervals, using a reference time zone to create a set of snapshot graphs. In a snapshot graph of a blockchain {transaction} network, a node represents a {participant (traders, exchanges, liquidity pools, etc.)}, whereas a {directed} edge denotes a financial transaction {involving the transfer of assets from one participant to the other}. Next, we define {\textsf InnerCore}, {\textsf InnerCore} expansion, and {\textsf InnerCore} decay on the snapshot graphs. {\textsf InnerCore} helps us eliminate unimportant edges and nodes (e.g.,  addresses trading small amounts).  We then compute daily temporal {\textsf InnerCore} expansion and decay measures to identify significant days and trends for further investigation (\S\ref{sec:methdecayexpansion}, \S\ref{sec:methpatterns}). Subsequently, centered-motif analysis and {\textsf NF-IAF} score percentile ranking is employed to capture anomalous addresses of market manipulator traders (\S\ref{sec:methmotif}). 
\subsection{InnerCore of a Graph}
\label{sec:methinnercore}
{Consider the weighted, directed multi-graph defined in Section~\ref{sec:prob}.} We define data depth of a node $v \in \V_t$ as the degree of ``outlyingness" of the node properties in relation to the origin $\mathbf{0}$.  We use In-Degree, Out-Degree, In-Strength, and Out-Strength as node properties (defined in Table \ref{tab:node_property_functions}) to compute the {\textsf InnerCore} of a snapshot graph (\S\ref{sec:methinnercore}), as these node features can be defined easily for a weighted, directed, multi-graph.  

We define the {\textsf InnerCore} of $G$ as the set of nodes $\V^{inner}$ whose data depth, relative to themselves, is less than an $\epsilon$ value. We set $\epsilon$ to a small value, and iteratively recompute the depth of each node as we remove nodes whose data depth is greater than $\epsilon$ in each iteration. This process continues until no more nodes can be removed. The resulting set of nodes is the {\textsf InnerCore}. %of the graph. 
% The {\textsf InnerCore} computation is detailed in Algorithm~\ref{alg:innercore} in the Appendix.%

\SetKwInput{KwInput}{Input}
\SetKwInput{KwOutput}{Output}
\SetKwRepeat{Do}{do}{while}

\begin{algorithm}[tb!]
\footnotesize
\KwInput{Directed, weighted, multigraph $G(V,E,w)$,\\
Set of node property functions $p_1, ..., p_n \in P$,\\%,
Data depth threshold $\epsilon$}
\KwOutput{InnerCore $V^{inner}$}
\tcp{Compute feature matrix}
$F = [f_1, ..., f_n] = \forall p_i \in P: f_i = p_i(v, G), \forall v \in V$\label{alg:line1}\; %\tcp{initial feature matrix}\label{alg:line1}
$\Sigma_F^{-1}$ = cov$(F)^{-1}$; \tcp{compute only once}\label{alg:line2}
\tcp{Compute initial depth values}
$z = [z_1, ..., z_n] = \forall v_i \in V: z_i = [1+(F_{i,*})'\Sigma^{-1}_F(F_{i,*})]^{-1}$\label{alg:line3}\;
 \Do{$\exists z_i: (z_i \geq \epsilon) \wedge (v_i \in V)$}{
   \ForEach{$z_i \geq \epsilon$}{
   $\V = \V \setminus \{v_i\}$\label{alg:line14}\;
   }
   \tcp{recompute node properties}
   $F = \forall p_i \in P: p_i(v, G), \forall v \in V$\label{alg:line16}\;
   \tcp{recompute depth}
   $z_i = [1+(F_{i,*})'\Sigma^{-1}_F(F_{i,*})]^{-1}, \forall v_i \in V$\label{alg:line17}\;
 }
\KwRet{$\V$ \tcp{as InnerCore $V^{inner}$}}\label{alg:line21}
\caption{\small{\textsf InnerCore} Discovery}
\label{alg:innercore}
\end{algorithm}

Algorithm~\ref{alg:innercore} computes a feature matrix $F$ based on each node property function in line~\ref{alg:line1}. {
In particular, edge weight is used for computing Strength, In-Strength, and Out-Strength node property functions, where the numerical values of all incident edges to a node irrespective of direction, inbound to a node, and outbound from a node, respectively, are aggregated. For example, if we have a network $A \xrightarrow[]{\text{10}} B \xleftarrow[]{\text{5}} C$  , the In-Strength node property function will return 15 for node B. 
%Each selected node property function sums its corresponding node attribute and generates a separate column in the matrix with the summation results for each node. For instance, from Table~\ref{tab:node_property_functions}, Strength, In-Strength, and Out-Strength node property functions sums the weights of all incident edges to a node irrespective of direction, inbound to an node, and outbound from a node, respectively.
} The feature matrix $F$ is used to compute the inverse covariance matrix $\Sigma_F$ in line~\ref{alg:line2}, which will be utilized for future data depth calculations. The initial depth of each node is determined using the Mahalanobis depth with respect to the origin at line~\ref{alg:line3}. Nodes with a depth greater than or equal to input $\epsilon$ are removed from the node-set $\V$ at line~\ref{alg:line14}. Once one batch of node removals has been performed, the feature matrix and depth values are re-evaluated in lines~\ref{alg:line16}--\ref{alg:line17}. If any remaining nodes still have a depth greater than or equal to $\epsilon$, the next batch is initiated at the same $\epsilon$ level. When there are no nodes left with a depth larger than $\epsilon$, the algorithm is considered complete, and the remaining nodes in $\V$ are returned as the {\textsf InnerCore}.

% \begin{table}
% \fontsize{9}{12}\selectfont
% %\vspace{-1mm}
% %\centering
% \footnotesize
% \caption{\small Example node property functions.}
% \label{tab:node_property_functions}
% \vspace{3mm}
% \begin{tabular}{ll}
% %\toprule
% Function & Definition\\
% \midrule
% $N(v)$ & neighbors of $v$ \\
% $N_{out}(v)$ & neighbors reachable with outgoing edges from $v$ \\
% $N_{in}(v)$ & neighbors reachable with incoming edges to $v$ \\
% $deg(v)$ & edges to/from $v$ (Degree) \\
% $deg_{out}(v)$ & outgoing edges from $v$ (Out-Degree) \\
% $deg_{in}(v)$ & incoming edges to $v$ (In-Degree) \\
% $S(v)$ & sum of edge weights incident to a node (Strength)\\
% $S_{out}(v)$ & sum of outgoing edge weights (Out-Strength)\\
% $S_{in}(v)$ & sum of incoming edge weights (In-Strength)\\
% % $\bigcirc(u, l)$ & undirected cycles of length $l$ that $u$ is part of \\
% % $\circlearrowright(u, l)$ & directed cycles of length $l$ that $u$ is part of \\
% % $t(u, l)$ & length $l$ timeframes that $u$ has edges in \\
% %\bottomrule
% \end{tabular}
% % \vspace{-3mm}
% \end{table}

\begin{table}
\fontsize{8}{10}\selectfont
\caption{\small Example node property functions.}
\label{tab:node_property_functions}
\begin{tabular}{ll}
\toprule
Function & Definition \\
\midrule
$N(v)$ & neighbors of $v$ \\
$N_{out}(v)$ & neighbors reachable with outgoing edges from $v$ \\
$N_{in}(v)$ & neighbors reachable with incoming edges to $v$ \\
$deg(v)$ & edges to/from $v$ (Degree) \\
$deg_{out}(v)$ & outgoing edges from $v$ (Out-Degree) \\
$deg_{in}(v)$ & incoming edges to $v$ (In-Degree) \\
$S(v)$ & sum of edge weights incident to a node (Strength)\\
$S_{out}(v)$ & sum of outgoing edge weights (Out-Strength)\\
$S_{in}(v)$ & sum of incoming edge weights (In-Strength)\\
\bottomrule
\end{tabular}
\end{table}

\medskip\noindent\textbf{InnerCore vs. Alphacore.} {\textsf InnerCore} discovery of a graph $G$ does not require a complete decomposition of all graph cores by varying $\epsilon$, as it is done in {\textsf AlphaCore}~\citep{victor2021alphacore}. Instead, we set an $\epsilon$ value (e.g., $\epsilon=0.1$) just once, and then use the value to iteratively prune nodes until all remaining nodes, relative to themselves, satisfy a data depth less than $\epsilon$. The  {\textsf InnerCore} approach is also different from graph-$k$-core decomposition~\citep{BatageljZ11}, where the outer cores are computed first before the higher $k$-core can be determined. As a result, {\textsf InnerCore} discovery is quite scalable and can be applied to very large graphs. Our experiments in \S\ref{sec:exp} reveal that {\textsf InnerCore} discovery has a running time that is only one-tenth of that required for {\textsf AlphaCore} decomposition.

\medskip\noindent\textbf{Scalability}. Computing the {\textsf InnerCore} requires performing Cholesky decomposition on the covariance matrix at line~\ref{alg:line2} once, which has time complexity $O(d^3)$ for $d$ features. Node features need to be recomputed at each iteration of the while loop with a cost of $O(|\V|\times deg)$, where $deg$ is the average degree in the graph. There are at most $|\V|$ iterations (number of nodes). In the worst case, the total time complexity is $O(d^3 + |\V|\times deg \times |\V|)$. However, since the neighborhood of a node can be sparse, the value of $deg$ is small. Moreover, since multiple nodes are removed in batches, the number of iterations is much smaller than $|\V|$. For example, in a network with approximately 480,000 nodes and 1 million edges (\S\ref{sec:datasets}), only 4 iterations on average are needed for an $\epsilon$ = 0.1.

\subsection{InnerCore Expansion and Decay}
\label{sec:methdecayexpansion}

By analyzing how a temporal graph expands and shrinks in relation to %the 
entry and exit of nodes on a daily basis, we %can 
gain valuable insights into market sentiment. %Specifically, 
We define the influential nodes of a graph as its {\textsf InnerCore} nodes (i.e., $\V_t^{inner}$). %Next, we investigate how the {\textsf InnerCore} of a network expands and decays on a given day compared to previous days. 
We propose two measures to quantify the activity of influential nodes in the network: expansion and decay. {\em Expansion} counts the number of new influential nodes on day $t$ that were not influential in the preceding $i$ days, while {\em decay} quantifies the number of influential nodes from the previous $i$ days that are not present in the influential nodes of day $t$. {The goals of measuring {\textsf InnerCore} expansion and decay are two-fold: {\bf (1)} Correctly accentuate anomalous days to motivate further analysis using motifs and {\textsf NF-IAF} scores ranking; and {\bf (2)} accurately depict trends in the market to provide a sentiment indicator and explain mood. {\textsf InnerCore}, based on its output, isolates the key participants in the daily transaction network snapshot, whereas the expansion and decay measures provide a unique perspective on market trends and sentiment from the activity of key participants. As prefaced in \S\ref{sec:prob}, our {\textsf InnerCore} methodology focuses on detection rather than prediction, acknowledging the inherent unpredictability of malicious transactions originating from the external world.}

To this end, we first discover $\V_t^{inner}$ as the set of nodes in the {\textsf InnerCore} of the snapshot graph at timestamp $t$, and define $\V_{\cup(t-i)}^{inner}=\bigcup_i\V_{t-i}^{inner}$ as the union set of nodes in the {\textsf InnerCore} of snapshot graphs from timestamps $\{t-1,t-2,\ldots, t-i\}$ for $i\geq 1$. Next, we define the expansion and decay measures at timestamp $t$. %as follows:
 \begin{definition}[Expansion]
$\mathbb{E}_t=\left|\V^{inner}_t \setminus \V^{inner}_{\cup(t-i)}\right|.$
 \end{definition}

The expansion values have a range $[0,\infty)$, where a value $\geq$ 1 indicates the addition of new influential nodes in the {\textsf InnerCore}. 
\begin{definition}[Decay]
$\mathbb{D}_t=\left|\V^{inner}_{\cup(t-i)} \setminus \V^{inner}_{t}\right|.$
\end{definition}
The decay values have a range $[0,\infty)$; a value of 0 indicates that all {\textsf InnerCore} members from $\{t-1,t-2,\ldots, t-i\}$ are present in $t$. 

\begin{exam} [Expansion and Decay]
Suppose we have a temporal graph that produces two daily snapshot graphs at days $t$ and $t+1$. On day $t$, the InnerCore is composed of five nodes: $\V^{inner}_t=\{v_1, v_2, v_3, v_4, v_5\}$. On day $t+1$, the InnerCore has expanded to include eight nodes: $\V^{inner}_{t+1}=\{v_3, v_4, v_5, v_6, v_7, v_8, v_9, v_{10}\}$.

If we set $i=1$, we can calculate the expansion and decay measures for the day $t+1$ based on the previous day. In this case, the union of the InnerCores is $\V^{inner}_{\cup(t-i)}=\{v_1, v_2, v_3, v_4, v_5\}$. Therefore, we have:

The expansion measure $\mathbb{E}_{t+1}$ is equal to $\left|\{v_6, v_7, v_8, v_9, v_{10}\}\right|=5$. %which yields a value of 5.  
The decay measure $\mathbb{D}_{t+1}$ is equal to $\left|\{v_1,v_2\}\right|=2$. %which yields a value of 2.
\end{exam}

A substantial expansion measure observed on a particular day often indicates the presence of excessive buy or sell behavior from new traders entering the daily {\textsf InnerCore}.  Such behavior may arise either from a large group of traders acting in unison or from a selected group of traders whose significant transactions prompt other traders to follow a similar pattern. Consequently, heavy-buy or heavy-sell behaviors coincide on days characterized by considerable influxes of new traders entering the daily {\textsf InnerCore}. On the other hand, a substantial decay measure observed on a particular day often is reactionary in response to a significant change in the state of a currency caused by the transactions of key traders in the preceding days. Therefore, we suggest that days with significant expansion measures, followed by days with significant decay measures, as anomalies and prime candidates for detecting market manipulator addresses.
\begin{figure}
    \centering \includegraphics[width=1\linewidth]{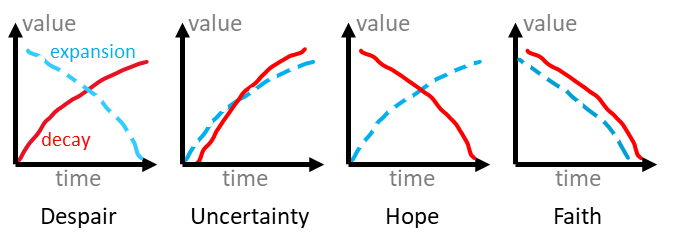}
    \vspace{-5mm}
    \caption{\small In a temporal graph (e.g., transaction network), changes in decay and expansion
    reflect varying levels of hope, despair, uncertainty, and faith in the asset being represented.
    }
    \label{fig:behavior}
\end{figure}

\spara{Parameters in Experimental Setup.}
In the context of {\textsf InnerCore} expansion and decay,
a greater $i$ (i.e., the history parameter from \S3.2) produces an averaging effect, coupled with the tendency to lower expansion and inflate decay.  Setting a specific $i$ value depends on the application. %in general, higher $i$ reduces fluctuations between expansion and decay of each day.  
%In our experiments, 
We use $i$ = 1 to improve the accentuation of expansion and decay in the {\textsf InnerCore} to better depict the shift in market sentiment during the days of significant events. %in our case studies.}  

In {\textsf InnerCore} decomposition, depth values range between $(0, 1]$; nodes with high property values (e.g., many transactions, higher transacted amounts) tend to have low depth, while nodes with low property values tend to have high depth~\citep{victor2021alphacore}. With data depth threshold $\epsilon=1$, all nodes will be returned as {\textsf InnerCore}  members; while for $\epsilon=0$, the empty set will be returned.  
Setting an appropriate $\epsilon$ depends on the desired size of the {\textsf InnerCore} returned specific to an application.  In our experiments, we set $\epsilon=0.1$ to ensure that the average number of nodes in each daily {\textsf InnerCore} is above 150.

\subsection{Behavioral Patterns in Temporal Networks}
\label{sec:methpatterns}
Temporal networks, including blockchain networks, exhibit continuous evolution and can experience notable shifts in user sentiment and node activity triggered by technological advancements and significant events, sometimes occurring within fewer days.

By utilizing expansion and decay, we have identified four behavioral patterns that {provide sentiment indication} and capture node activity. These patterns serve as the foundation for network analysis in our experiments detailed in \S\ref{sec:exp}. Figure~\ref{fig:behavior} illustrates the expansion and decay values for each pattern. To gain a better understanding of these patterns, particularly when examining the temporal graph of a financial network such as the Ethereum transaction network, it is helpful to consider the network's underlying transaction semantics.
\begin{itemize}[leftmargin=.1in]
    \item The {\em Despair} pattern is characterized by a reduction in expansion and an increase in decay, implying that previously influential nodes are leaving the network, while the {\textsf InnerCore} is shrinking due to a decrease in the number of new influential nodes.
    \item The {\em Uncertainty} pattern is distinguished by an increase in both expansion and decay. This is primarily due to the influx of many new traders into the network who do not remain active for a significant period of time.
    \item The {\em Hope} pattern is characterized by a reduction in decay and an increase in expansion, indicating the presence of many newcomers to the network who remain active within the network.
    \item The {\em Faith} pattern is identified by a decrease in both decay and expansion, which initially suggests a state of confusion. On the positive side, nodes, such as traders, may have faith in the network's ability to withstand a catastrophic event, as demonstrated in the LunaTerra case in our experimental results. On the negative side, it may indicate a sense of hopelessness as traders may hold onto their assets without engaging in transactions or exiting the system altogether.
\end{itemize}
\subsection{Motif Analysis in InnerCore}
\label{sec:methmotif}
Our rationale behind using motif analysis in conjunction with {\textsf InnerCore} is to accurately discover larger and potentially influential players in the daily network, referred to as market manipulators.  The structure of a motif defines a behavior of interest and its existence in a network indicates the presence of such behavior.  
\begin{figure}
    \centering \includegraphics[width=1\linewidth]{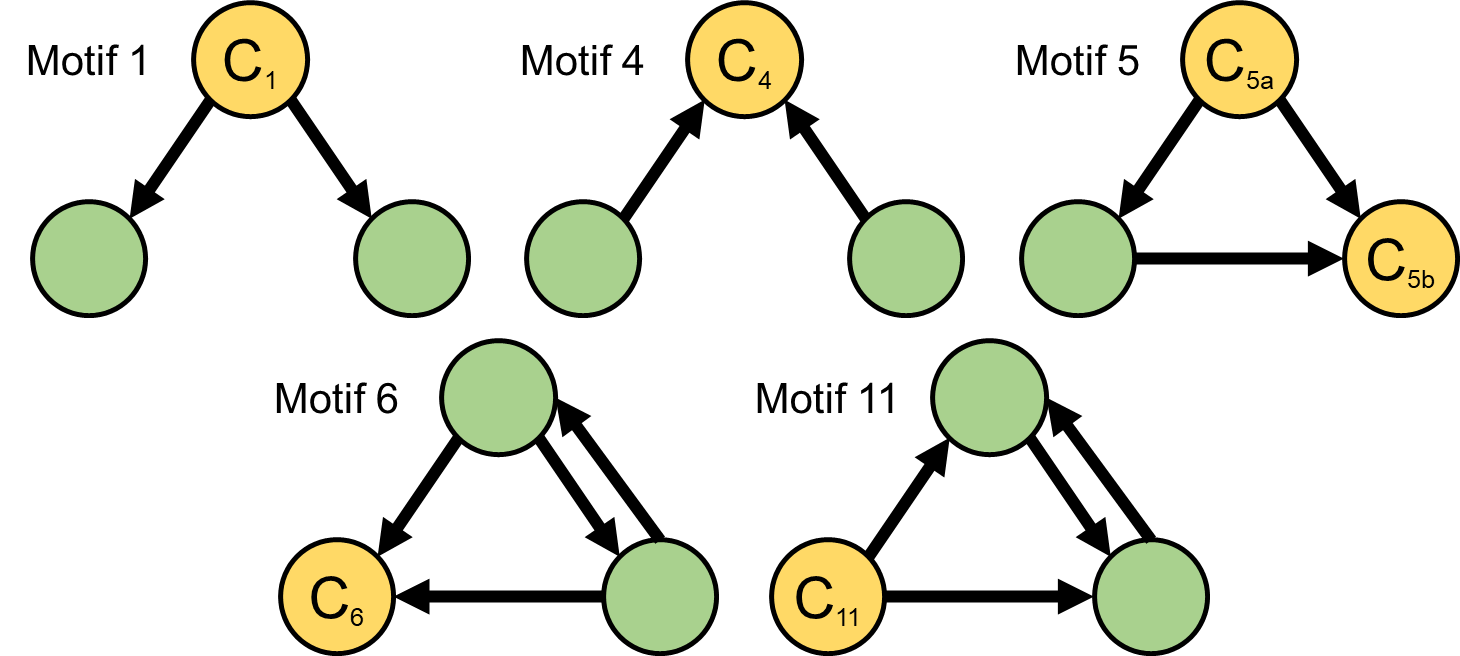}
    \vspace{-3mm}
    \caption{\small Five 3-node motifs exhibiting buy and sell behaviors.  Nodes labeled C denote the center where a center with an in-degree = 2 indicates buy behavior and an out-degree = 2 indicates sell behavior. Out of the 16 connected 3-node motifs (see Figure 1B in \citet{milo2002network}), only the five given above (motifs 1, 4, 5, 6, and 11) contain a center node. 
    \label{fig:motifs}}
    % \vspace{-5.5mm}
\end{figure}

Motif analysis has been a popular tool to identify subgraph patterns and the addresses involved in them \citep{LeeKGOL20,bailey2009meme,zhang2012extracting,paranjape2017motifs,milo2002network}. We have decided to use three-node motifs since they can be identified more quickly than higher-order motifs, while still capturing the direct buying or selling behavior between addresses.  Our decision is consistent with previous research on temporal motifs~\citep{paranjape2017motifs}.  

\medskip
\noindent\textbf{Scalability}. The fastest triangular motif discovery algorithm has time complexity $O(|\V^{inner}|^\omega)$, where $\omega < 2.376$ is the fast matrix product exponent~\citep{latapy2008main,coppersmith1987matrix}. The number of nodes in the {\textsf InnerCore} is denoted by $|\V^{inner}|$. We demonstrate in \S\ref{sec:exp} that triangular motif discovery on {\textsf InnerCore}s has low time costs because of the relatively small size of daily networks' {\textsf InnerCore}s.
In particular, we consider a simpler implementation of triangular motif discovery,
where for each node we explore its local neighborhood. For every triple consisting of the current node and its two neighbors, we verify if a motif can be formed.  
The time complexity of our approach is 
$O\left(|\V^{inner}| \times {nbr \choose 2}\right)$,
where $nbr$ denotes the maximum number of neighbors per node. The daily temporal {\textsf InnerCore} networks from our Ethereum stablecoin dataset have, on average, 180 nodes, with each node having 11 neighbors on average (max. number of neighbors of a node = 134).
In contrast, the entire daily temporal Ethereum stablecoin networks have, on average, 89,500 nodes and though each node has only 3 neighbors on average, the maximum number of neighbors per node is 69,381. This explains why our triangular motif discovery method is quite efficient on the {\textsf InnerCore} networks as opposed to on entire daily temporal graphs. 

%Innercore max N: 134    InnerCore average |V| = 179.31
%Whole network max N: 69381    whole network average |V| = 89464.21

We define the center of each 3-node motif as a node that either receives incoming edges from the two other nodes (buy behavior) or delivers outgoing edges to two other nodes (sell behavior). This definition ensures that motif centers 
exhibit only buy or sell behavior, 
and they do not act as intermediary nodes between the other two nodes in a motif.

Out of the 16 connected three-node motifs (see Figure 1B in \citet{milo2002network}), only five of them contain a center node (Figure~\ref{fig:motifs}).
We identify all instances of these five motifs and their centers from our daily networks' {\textsf InnerCore}s. 
Finally, we utilize the well-known {\textsf TF-IDF} measure from information retrieval~\citep{salton1988term} to rank the discovered center nodes. 
{\textsf TF-IDF} is a statistical measure to reflect the relevance of a word in a collection of documents. In our setting, we treat each discovered center address as a word and daily instances of each motif as a collection of documents to propose a novel node relevance score for temporal graphs: {\textsf NF-IAF}.   

Formally, let $M={m_1,m_4,m_{5},m_{6},m_{11}}$ be the set of five motifs of interest, and let $T={t_1,t_2,\dots,t_n}$ be the set of $n$ days under consideration. 
For each $m_i \in M$ and $t_j \in T$, let ${c(v,m_i,t_j)}$ denote the number of occurrences of node $v\in \V^{inner}$ in all instances of motif $m_i$ on day $t_j$.
For all $v\in \V^{inner}$, $m_i\in M$, and $t_j\in T$, we define the node frequency ({\textsf NF}) and inverse-appearance frequency ({\textsf IAF}) as follows:
\begin{definition}[Node Frequency]
We define the node frequency of node $v$ for motif $m_i$ on day $t_j$  as 
$$NF(v,m_i,t_j)=  \frac{c(v,m_i,t_j)}{\sum\limits_{v \in \V_j^{inner}}{c(v,m_i,t_j)}}.$$ % 
\end{definition}
The {\textsf NF} measures how frequently a particular node occurs in a specific motif on a specific day relative to the total number of occurrences of all nodes in that motif on that day. 
\begin{definition}[Inverse Appearance Frequency]
We define the inverse appearance frequency of node $v$ for motif $m_i$  as 
$$IAF(v,m_i) = \log\frac{|T|}{df(v,m_i)}$$
where $|T|$ is the total number of days in the dataset, and $df(v,m_i)$ is defined as the number of days $t_j\in T$ where $c(v,m_i,t_j)>0$.
\end{definition}
The {\textsf IAF} measures the importance of a node by how frequently it appears across all days for a motif.  If a node appears in many days for a motif, its {\textsf IAF} will be low, indicating that it is not very informative. On the other hand, if a node appears in only a few days for a motif, its {\textsf IAF} will be high, indicating that it is a rare and potentially important node.
\begin{definition}[NF-IAF Score]
The {\textsf NF-IAF} score of node $v$ for motif $m_i$ on day $t_j$ is given as 
$$NF{\text-}IAF(v,m_i,t_t) = NF(v,m_i,t_j) \times IAF(v,m_i).$$
\end{definition}
A greater {\textsf NF-IAF} score of a center node on a particular day  
indicates greater relevance between that node and the behavior associated with the motif type.  Therefore, a node corresponding to a motif center on a particular day with a high {\textsf NF-IAF} score has an increased likelihood that it has more influence on the network on that day, while a lower {\textsf NF-IAF} score indicates the opposite.  

\input{sections/052_tfidf_example.tex}

%% file: sections/052_tfidf_example.tex
\begin{exam}
 Table~\ref{tab:tfidfex_results_combined} shows the number of occurrences of three nodes over three days for motifs $m_4$ and $m_5$. For example, to compute ${NF}(v_1, m_4, t_1) = \frac{5}{5+15+0} = 0.25$, we divide the number of times $v_1$ appears in instances of $m_4$ on day $t_1$, by the total number of occurrences of all nodes in instances of $m_4$ on day $t_1$. Similarly, we compute $IAF(v_1,m_4,t_1) = log(\frac{3}{3}) = log(1) = 0$ as $v_1$ appears in all three days for $m_4$. Thus, $NF{\text -}IAF(v_1,m_4,t_1)=0.25\times 0 = 0$. The resulting {\sf NF-IAF} score for each node, motif, and day combination is given in the right panel 
 of Table~\ref{tab:tfidfex_results_combined}.

\iffalse
\begin{table}[tb!]
\caption{Number of occurrences of nodes $v_1,v_2$, and $v_3$ across three days $t_1,t_2$, and $t_3$ in instances of motifs $m_4$ and $m_5$. $v_3$ does not appear for motif $m_4$ on any day, whereas $v_1$ does not appear on days $t_1$ and $t_2$ for motif $m_5$.}
\label{tab:tfidfex}
\footnotesize
\begin{tabular}{lcccccc}
 & \multicolumn{3}{c}{$m_4$} & \multicolumn{3}{c}{$m_5$} \\
 \toprule
 node    & $t_1$      & $t_2$      & $t_3$     & $t_1$      & $t_2$      & $t_3$     \\
   \cmidrule(lr){2-4}   \cmidrule(lr){5-7}
$v_1$   & 5       & 4       & 3      & 0       & 0       & 15      \\
$v_2$   & 15      & 12      & 9      & 4       & 7       & 9      \\
$v_3$   & 0       & 0       & 0      & 21      & 23      & 35  \\
%\bottomrule
\end{tabular}
\end{table}
%
\begin{table}[tb!]
\caption{{\sf NF-IAF} scores of nodes based on the motif information in Table~\ref{tab:tfidfex}. When ranking nodes with their {\sf NF-IAF}, we find $v_1$ on $t_3$ as the highest ranked node for motif $m_5$.}
\label{tab:tfidfex_results}
\centering
\footnotesize
\begin{tabular}{lcccccc}
& \multicolumn{3}{c}{$m_4$} & \multicolumn{3}{c}{$m_5$} \\
\toprule
node & $t_1$ & $t_2$ & $t_3$ & $t_1$ & $t_2$ & $t_3$ \\
\cmidrule(lr){2-4} \cmidrule(lr){5-7}
$v_1$ & 0.25 & 0.25 & 0.25 & 0 & 0  & 0.92 \\
$v_2$ & 0.75 & 0.75 & 0.75 & 0.16 & 0.23 & 0.26 \\
$v_3$ & 0 & 0 & 0 & 0.84 & 0.74 & 0.71 \\
%\bottomrule
\end{tabular}
\end{table}
\fi

\begin{table}[tb!]
\caption{\small Occurrences and NF-IAF scores of nodes $v_1,v_2$, and $v_3$ across three days $t_1,t_2$, and $t_3$ in instances of motifs $m_4$ and $m_5$. $v_3$ does not appear for motif $m_4$ on any day, whereas $v_1$ does not appear on days $t_1$ and $t_2$ for motif $m_5$.}
\label{tab:tfidfex_results_combined}
\centering
\footnotesize
\begin{tabular}{lcccccccccccc}
& \multicolumn{6}{c}{Occurrence}  & \multicolumn{6}{c}{NF-IAF Score} \\
\cmidrule(lr){2-7}  \cmidrule(lr){8-13} 
& \multicolumn{3}{c}{$m_4$} & \multicolumn{3}{c}{$m_5$} & \multicolumn{3}{c}{$m_4$} & \multicolumn{3}{c}{$m_5$} \\
\cmidrule(lr){2-4}\cmidrule(lr){5-7} \cmidrule(lr){8-10}\cmidrule(lr){11-13}
node&$t_1$&$t_2$&$t_3$&$t_1$&$t_2$&$t_3$&$t_1$&$t_2$&$t_3$&$t_1$&$t_2$&$t_3$\\
\cmidrule(lr){2-4}\cmidrule(lr){5-7} \cmidrule(lr){8-10}\cmidrule(lr){11-13}
$v_1$&5&4&3&25&0&0& 0 & 0 & 0 & 0.48 & 0  & 0 \\
$v_2$&15&0&9&0&7&13& 0.13 & 0 & 0.13 & 0 & 0.04 & 0.05 \\
$v_3$&0&0&0&0&23&35& 0 & 0 & 0 & 0 & 0.14 & 0.13 \\

%\bottomrule
\end{tabular}
% \vspace{-3mm}
\end{table}

 \end{exam}

%% file: sections/06_experiments.tex
\section{Experimental Results}
\label{sec:exp}
In this section, we first describe three large temporal blockchain graphs that we use to answer our research questions (\S\ref{sec:prob}). Next, we analyze the scalability of {\textsf InnerCore} discovery and centered-motif analysis on these graphs. Upon demonstrating our scalability results, we illustrate how our methods provide predictive insights into anomalies stemming from external events and identify the addresses that played a significant role in such events. Our code and datasets are available at {\url{https://github.com/JZ-FSDev/InnerCore}}.
\vspace{-2mm}
\subsection{Environment Setup}
\subsubsection{Datasets}
\label{sec:datasets}
Our experiments investigate the Ethereum transaction network and Ethereum stablecoin networks across three recent real-world events: the LunaTerra collapse, Ethereum's transition to Proof-of-Stake, and USDC's temporary peg loss. {For each of our experiments, we construct a transaction network from the following datasets.}

\noindent\textbf{Ethereum Stablecoin Transaction Networks}. We retrieve transaction data for the top five stablecoins based on market capitalization (UST, USDC, DAI, UST, PAX) and WLUNA from the Chartalist repository~\citep{Chartalist2022}. The data pertains only to transactions conducted on the Ethereum blockchain; each transaction in the data set corresponds to a transfer of the asset indicated by the contract address.
However, the UST collapse event that we are studying involved another blockchain called Terra with its own network, and the cryptocurrency called Luna, acting as a parallel to ether on Ethereum. Terra issued a stablecoin named UST (also known as TerraUSD), which offered high-interest rates to lenders and was pegged to the value of \$USD1. Additionally, Terra's owners created an ERC-20 version UST on the Ethereum blockchain and a Wrapped LUNA (WLUNA) token was established to trade Luna tokens on Ethereum. In May 2022, the Terra blockchain and its cryptocurrency Luna collapsed, owing to TerraUSD loans that could not be repaid. A Luna coin that was valued at \$USD116 in April plummeted to a fraction of a penny during the collapse\footnote{{\tiny \url{https://coinmarketcap.com/currencies/wrapped-luna-token/}}}. This resulted in a loss of confidence in both WLUNA and UST on Ethereum. On May 9th, 2022, UST lost its \$USD1 peg and fell as low as 35 cents\footnote{{\tiny \url{https://coinmarketcap.com/currencies/terrausd/}}}.  The Ethereum Stablecoin dataset covers the period from April 1st, 2022, to November 1st, 2022, spanning about one month before the crash to six months after the crash. %In addition to the transactions, 
{We construct a transaction network consisting of UST, USDC, DAI, UST, PAX, and WLUNA transactions for \S\ref{sec:exp1} between this period.}
We also use the address labels dataset from~\citet{Chartalist2022} where labels of 296 addresses from 149 centralized and decentralized Ethereum exchange addresses are listed publicly {to distinguish unique exchange addresses}.

In March 2023, Silicon Valley Bank, holding over 3 billion of Circle’s collateralized reserves collapsed abruptly, causing a mass liquidation of USDC from traders. Consequently, on March 11th, 2023, Circle's USDC temporarily lost its \$USD1 peg, dropping to an all-time low of 87 cents.  The USDC dataset covers the period from February 25th, 2023, to March 23, 2023, spanning approximately two weeks before and after the peg loss. {We use a transaction network consisting of only USDC transactions for \S\ref{sec:exp3}.}

\noindent\textbf{Ethereum Transaction Network}. We collected ether transactions from the Ethereum blockchain for the period between August 21st and October 1st, 2022. On an average day during this period, there were 480,000 addresses, with approximately 1 million edges connecting them. Ether is a type of cryptocurrency, similar to bitcoin, and its value can be converted to various fiat currencies such as USD and JPY. 
% \textcolor{red}{The nodes on the graph could represent traders who anticipate future price increases for ether or traders who engage in buying and selling goods and services.} 
Ethereum changed its block creation process during this time, moving from the costly Proof-of-Work method to the more efficient Proof-of-Stake algorithm in two phases on September 9th and 15th, 2022.

\subsubsection{Competitors} We compare {\textsf InnerCore} with two baselines: {\textsf AlphaCore}~\citet{victor2021alphacore} and graph-$k$-core~\citep{BatageljZ11}. We refer to \S\ref{sec:methinnercore}, \textbf{InnerCore vs. Alphacore} for their differences. Additionally, we compare against {\textsf Scalable Change Point Detection (SCPD)} \citep{huang2023fast}, state-of-the-art attributed change detection method in dynamic graphs.% It is a novel spectral method to identify anomalies from a set of graph snapshots.
\begin{figure}
  \centering      \includegraphics[width=0.45\textwidth]{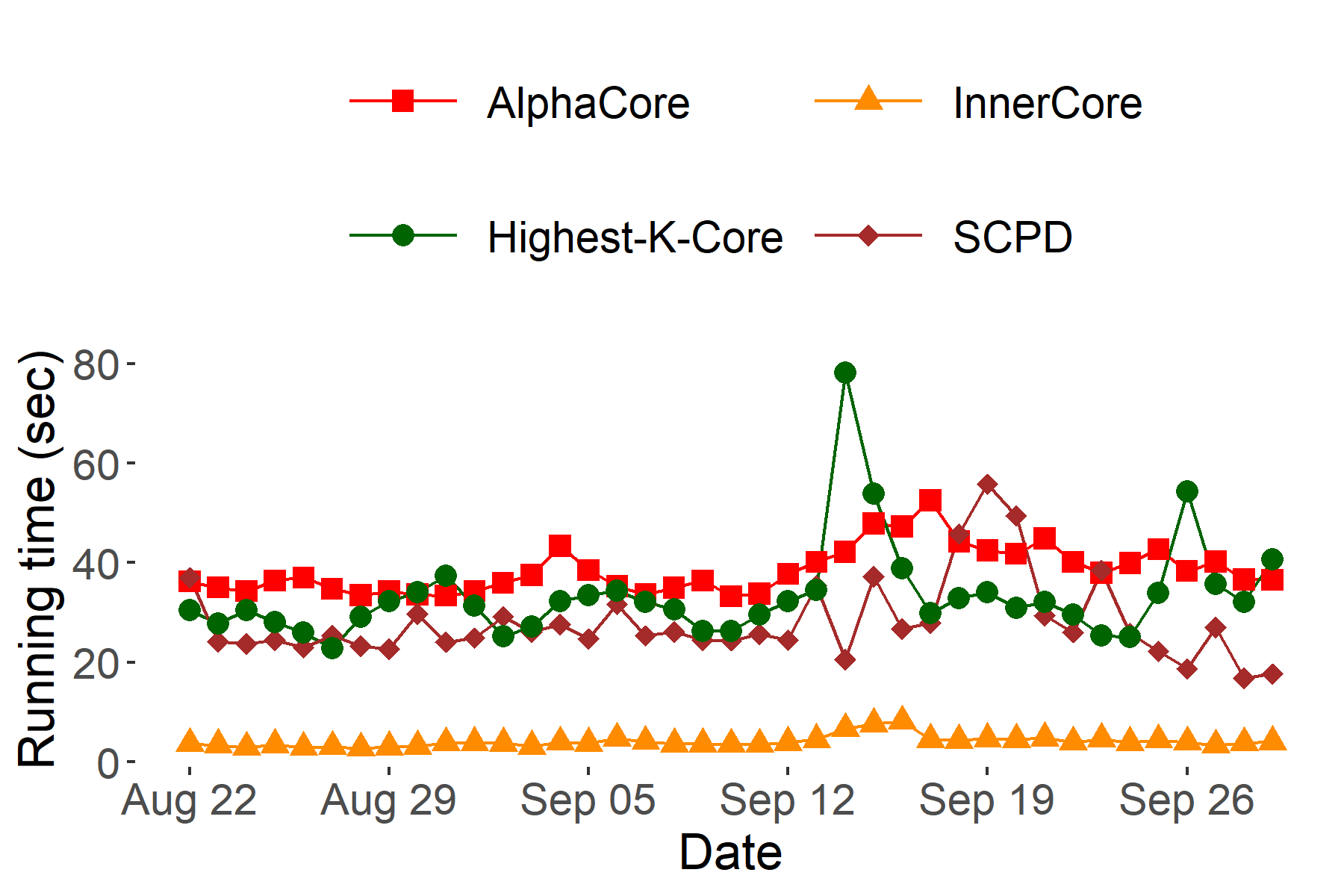}
  \vspace{-3mm}
  \caption{\small Comparison between running times of {\textsf AlphaCore} with the starting $\epsilon=1.0$ and stepsize $s=0.1$, {\textsf InnerCore} with $\epsilon=0.1$ on daily Ethereum transaction networks to return the {\textsf InnerCore} of depth $<$ 0.1. %, and SCPD's DOS computation.  
  An average of approximately 480,000 nodes (addresses) and 1 million edges (transactions) exist in each network. The average computation time is 4.06 seconds (max 8.1s), which is approximately 0.10 times the average computation time of {\textsf AlphaCore}, 0.12 times the average computation time of the highest graph $k$-core, and 0.14 times the average computation time of {\textsf SCPD}. %'s DOS.  
  }
  \label{fig:time}
  % \vspace{-12px}
\end{figure}

\vspace{-2mm}
 \subsection{Scalability Analysis}

\noindent\textbf{System Specifications}.
The machine used for experiments is an Intel Core i7-8700K CPU @ 3.70GHz processor, 32.0GB RAM, Windows10 OS, and GeForce GTX1070 GPU.  A combination of Python and R was used for coding.

%\medskip
\noindent\textbf{InnerCore Discovery}.
Since we are interested in directly finding the {\textsf InnerCore}, compared to {\textsf AlphaCore} decomposition~\citep{victor2021alphacore}, {\textsf InnerCore} discovery method (\S 3.1) does not associate different $\epsilon$ values to intermediate cores generated in an iterative stepwise fashion.  Instead, a fixed threshold $\epsilon$, or upper bound for depth, is set and all nodes with a depth greater than $\epsilon$ are pruned repetitively until all remaining nodes relative to each other in the resulting network have a depth  $< \epsilon$. This allows {\textsf InnerCore} discovery to run approximately 1/stepsize times faster than {\textsf AlphaCore} decomposition since the computations of all intermediate cores are skipped.
As depicted in Figure  \ref{fig:time}, the average running time for {\textsf InnerCore} discovery is only 4.06 seconds on graphs with approximately 480,000 nodes and 1 million edges. Furthermore, {\textsf InnerCore} discovery has a running time of only one-tenth of that for {\textsf AlphaCore} decomposition.

Due to the need for graph-$k$-core to repetitively iterate over all remaining nodes with each peeling until the highest $k$-core remains, we find {\textsf InnerCore} to be nearly 8x faster on each daily graph snapshot.

{\textsf SCPD} is state-of-the-art method to identify anomalies from attributed graph snapshots~\citep{huang2023fast}.  Due to its spectral approach, we find it slower:
%Similarly, we utilize {\textsf InnerCore} expansion and decay (\S 3.2) to identify anomalies from daily temporal networks.  To this end, we compare the state-of-the-art SCPD method to our {\textsf InnerCore} expansion and decay method.  Note {\textsf InnerCore} discovery is the precursor required to plot {\textsf InnerCore} expansion and decay measures for identification of anomalous days as the density of states (DOS) embedding computation is the precursor for SCPD~\cite{huang2023fast} to detect anomalies.  Thus, we timed the DOS embedding computation of each daily Ethereum network snapshot and observe from Figure \ref{fig:time} that compared to SCPD's DOS embedding computation, 
{\textsf InnerCore} discovery runs nearly 7x faster on each daily graph snapshot, which demonstrates the scalability of our solution.

\smallskip
\noindent\textbf{Three-Node Motifs Counting}.  
Instead of conducting motif analysis on all nodes, our approach utilizes the {\textsf InnerCore}. By focusing on this core subset of nodes, we are able to reduce the number of nodes in a daily network consisting of approximately 480,000 nodes and 1 million edges to an induced subgraph of roughly 300 nodes and 90,000 edges (counting multi-edges), resulting in a more manageable and efficient approach.
Although centered motif counting on each snapshot graph takes $>$ 1 day to complete, motif counting inside {\textsf InnerCore}  significantly improves the processing speed, requiring only $<$ 20 secs to complete, which illustrates our scalability.
\vspace{-2mm}
\subsection{Effectiveness Analysis}
\subsubsection{Experiment 1: The Collapse of LunaTerra}
\label{sec:exp1}
Stablecoins are meant to be a safe house as they are generally pegged to and maintain a 1:1 ratio with a fiat currency, resisting the volatility associated with other popular cryptocurrencies. Commonly, traders keep blockchain assets not needed for immediate use in a transaction as a stablecoin, analogous to people keeping extra money in a bank.  For this reason, The LunaTerra collapse was a historic event in the decentralized financial space as it questioned traders' trust in cryptocurrencies; if even stablecoins are susceptible to collapse, then is any cryptocurrency truly safe?

\begin{figure}
  \centering      \includegraphics[width=0.45\textwidth]{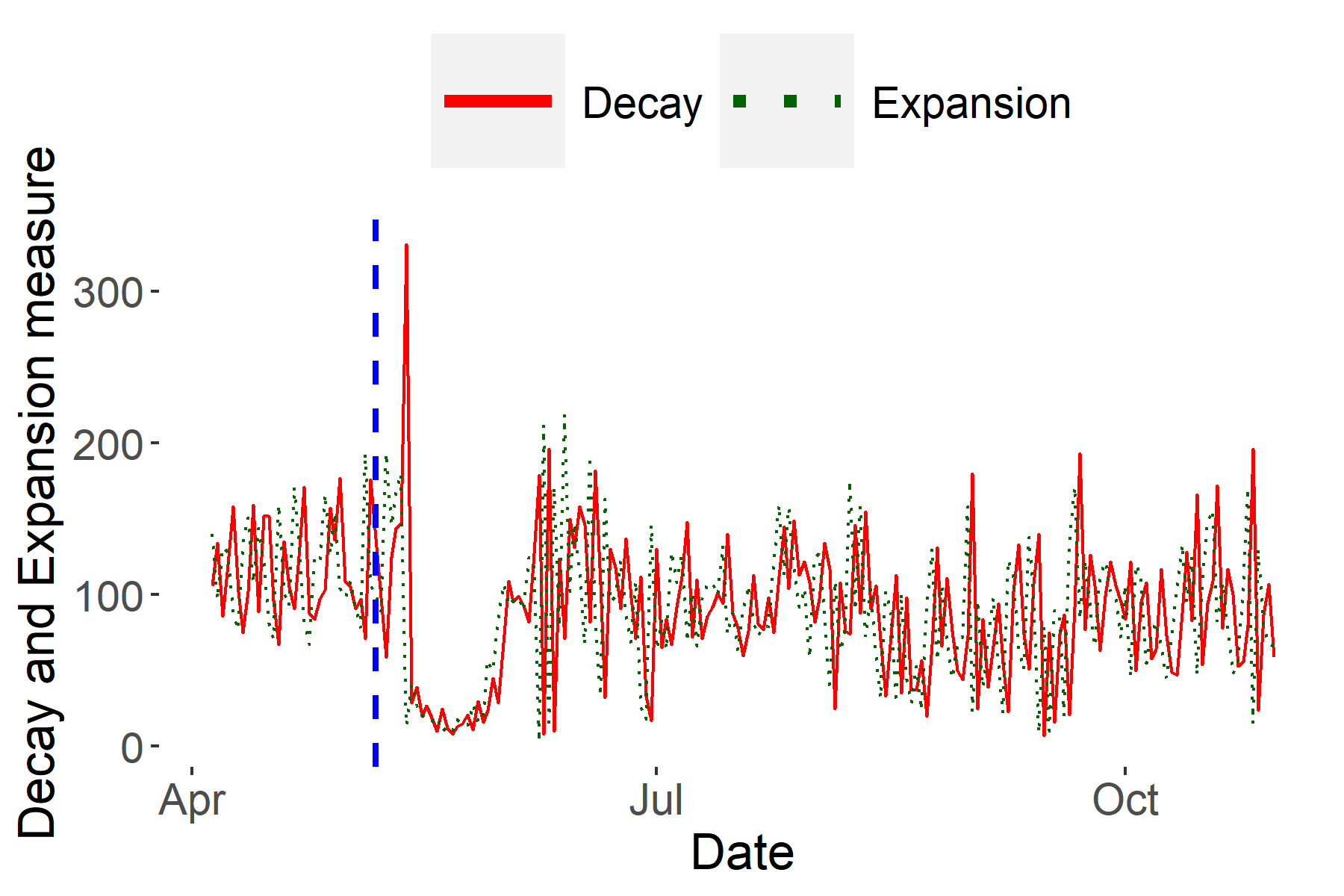}
  \vspace{-3mm}
  \caption{\small Stablecoin decay and expansion measures. On May 8 (shown with the vertical blue line), UST loses its \$1 peg and falls to as low as 35 cents.}
  \label{fig:stablecoinDecayExapansion}
  % \vspace{-5mm}
\end{figure}

\begin{figure}
  \centering      \includegraphics[width=0.45\textwidth]{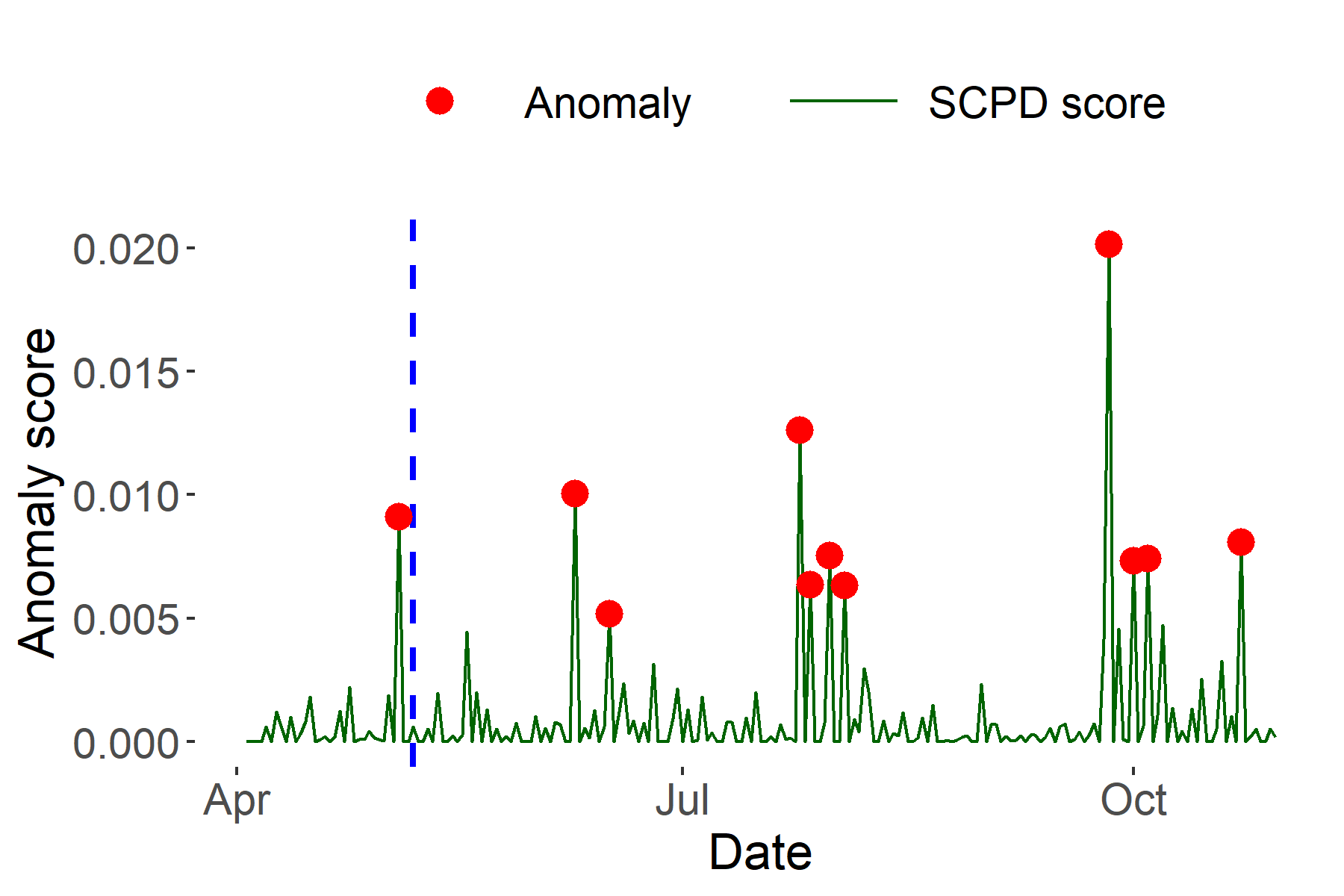}
  \vspace{-3mm}
  \caption{\small Stablecoin anomalous days identified by {\textsf SCPD}. Unlike decay and expansion measures by {\textsf InnerCore}, {\textsf SCPD} less accentuates the critical event of UST's peg loss in Ethereum stablecoin networks, compared to other anomalies that occurred between Apr 3 to Oct 30, 2022.}
  \label{fig:stablecoinSCPD}
  % \vspace{-5mm}
\end{figure}

\smallskip
\noindent\textbf{Behavioral Patterns via Expansion and Decay.}
First, we analyze this event from the perspective of traders' market sentiment %in the stablecoin network.  Specifically, we examine behavioral patterns in the 
via expansion and decay measures of the temporal stablecoin network for the days surrounding the collapse.  %From 
In Figure~\ref{fig:stablecoinDecayExapansion}, %we observe that 
four days after the collapse unfolded, on May 13, 2022, there was a substantial increase in decay and a decrease in expansion: a prime indicator of the {\em despair} behavioral pattern (\S 3.3).  We can infer from this signal that a large majority of regular traders stopped trading by this time, either from the conversion or sale of any assets stored as UST out of the stablecoin ecosystem or simply due to uncertainty and inaction in response to the collapse.  Following this cue, for approximately two weeks afterward, we see a consistent behavioral pattern of {\em faith} characterized by low expansion and low decay.  During this period, few new traders entered or left the stablecoin network.  There was still faith in the remaining traders that perhaps a large stablecoin such as UST could rebound and restore its peg with USD and thus, they refrained from engaging in any transactions.  On the other hand, decay and expansion values also indicate a sign of hopelessness as the bulk of traders already exited the network since the first signal of despair.  We understand from this behavioral analysis that there is a delayed reaction from traders when a significant unannounced event occurs due to indecision, and there is a general trend of inactivity in the following period.

\smallskip
\noindent\textbf{Why is this e-crime?} We outline two reasons. \textbf{Dumping of UST:} On May 7th, large sums of UST were dumped, with 85 million UST swapped for 84.5 million USDC \citep{liu2023anatomy}. This massive dumping of UST contributed to its de-pegging and caused its value to drop significantly. \textbf{Concealing past failures:} The CEO of Terra, Do Kwon, was revealed to be a co-creator of the failed algorithmic stablecoin, Basis Cash \citep{impekoven2023central}. The concealment of such information about the project's founder could mislead traders and hide potential risks. 

\smallskip
\noindent\textbf{SCPD vs. InnerCore}. 
% \textcolor{red}{We compare our 
%  expansion and decay results against the effectiveness of {\textsf SCPD} on Ethereum stablecoin networks.}  
From Figure~\ref{fig:stablecoinSCPD}, we observe that {\textsf SCPD} less accentuates the critical event of UST's peg loss {and {\textsf InnerCore} more accurately depicts the impact of the collapse on the market relative to other days in the data time span.}  %in Ethereum stablecoin networks.  SCPD 
{\textsf SCPD} assigns an anomaly score to Sep 26 when USDC announced their plan to expand to five new blockchains\footnote{{\tiny \url{https://www.chartalist.org/eth/StablecoinAnalysis.html}}}, nearly two times as anomalous as the score assigned to May 4, the closest day to the LunaTerra collapse.  However, our Stablecoin decay and expansion measures in Figure~\ref{fig:stablecoinDecayExapansion} notably accentuate and emphasize the impact of UST's peg loss on the stablecoin ecosystem from the less impactful events occurring on other days. This accentuation is evident by the presence of a pronounced decay peak on May 13 followed by a period of approximately two weeks of consistently low decay and expansion measures before returning to more standard values seen in other days, clearly indicating a significant event had transpired.  This demonstrates that decay and expansion measures serve as a better indicator of the significance of an event on its corresponding network.

% \begin{figure}
%   \centering      \includegraphics[width=0.65\textwidth]{figs/lunacores}
%   \vspace{-3mm}
%   \caption{\small Sizes of stablecoin network cores during LunaTerra collapse. {\textsf InnerCore}s accentuates the collapse better than nodes found in the highest cores via graph-$k$-core decomposition.  
%   }
%   \label{fig:lunaTerraCoreSize}
%   \vspace{-4mm}
% \end{figure}

 %
\begin{table}[t]
  \centering
  \caption{\small Numbers of center addresses in motifs identified by our method (\S 3.4) that are known exchanges. The numbers represent the total counts per motif across all days.}
  \label{tab:exchanges}
  % \vspace{3mm}
  \small
  \begin{tabular}{ccc}
    %\toprule
     & \# Unique Addresses & \# Exchange Addresses\\
    \midrule
    Motif 1 & 1221 & 15\\
    Motif 4 & 1762 & 15\\
    Motif 5 & 1447 & 17\\
    Motif 6 & 1513 & 4\\
    Motif 11 & 939 & 11\\
  %\bottomrule
\end{tabular}
% \vspace{-3mm}
\end{table}
\begin{table}[t]
  \centering
  \caption{\small {{\textsf NF-IAF} score percentile ranks of {\textsf InnerCore} motif centers  matching highlighted addresses by Nansen.ai to have played key roles before (May 7), during (May 8), and after (May 9, 2022) the LunaTerra collapse. The percentile scores for individual addresses on a specific day of a particular motif center are determined relative to all addresses associated with the same motif center throughout all days in the data window. Motif centers $C_1$, $C_{5a}$, $C_{11}$ exhibit sell behavior, while motif centers $C_4$, $C_{5b}$, $C_6$ exhibit buy behaviour. Addresses with percentiles $\geq$ 90 across at least one motif center type (given in red color) are considered impactful on a given day. Dashes indicate absence of the address as the motif center.}}
  \label{tab:nansenAddresses}
  % \vspace{3mm}
  \small
  \begin{tabular}{ccccccc}
    \multicolumn{7}{c}{LunaTerra addresses on May 7} \\
    \hline
    Address/Motif Center & $C_1$ & $C_4$ & $C_{5a}$ & $C_{5b}$ & $C_6$ & $C_{11}$ \\
    \hline
{\sf \textcolor{gray}{Celsius}}		&	\textcolor{gray}{-}	&	\textcolor{gray}{81}	&	\textcolor{gray}{79}	&	\textcolor{gray}{-}	&	\textcolor{gray}{-}	&	\textcolor{gray}{-}	\\
{\sf \textcolor{gray}{hs0327.eth}}		&	\textcolor{gray}{30}	&	\textcolor{gray}{4}	&	\textcolor{gray}{28}	&	\textcolor{gray}{28}	&	\textcolor{gray}{4}	&	\textcolor{gray}{-}	\\
{\sf Smart LP: 0x413}		&	\textcolor{gray}{-}	&	\textcolor{gray}{69}	&	\textcolor{gray}{-}	&	\textcolor{gray}{-}	&	\textcolor{red}{95}	&	\textcolor{gray}{-}	\\
{\sf \textcolor{gray}{Token Millionaire 1}}		&	\textcolor{gray}{85}	&	\textcolor{gray}{81}	&	\textcolor{gray}{73}	&	\textcolor{gray}{-}	&	\textcolor{gray}{67}	&	\textcolor{gray}{89}	\\
{\sf Token Millionaire 2}		&	\textcolor{gray}{35}	&	\textcolor{red}{100}	&	\textcolor{red}{99}	&	\textcolor{gray}{-}	&	\textcolor{red}{99}	&	\textcolor{gray}{38}	\\
{\sf masknft.eth}		&	\textcolor{red}{97}	&	\textcolor{red}{94}	&	\textcolor{gray}{82}	&	\textcolor{gray}{-}	&	\textcolor{red}{93}	&	\textcolor{red}{92}	\\
{\sf \textcolor{gray}{Heavy Dex Trader}}		&	\textcolor{gray}{54}	&	\textcolor{gray}{17}	&	\textcolor{gray}{-}	&	\textcolor{gray}{-}	&	\textcolor{gray}{32}	&	\textcolor{gray}{-}	\\
{\sf Oapital}		&	\textcolor{red}{94}	&	\textcolor{gray}{83}	&	\textcolor{gray}{62}	&	\textcolor{gray}{62}	&	\textcolor{gray}{72}	&	\textcolor{red}{92}	\\
{\sf Hodlnaut}		&	\textcolor{gray}{40}	&	\textcolor{red}{99}	&	\textcolor{red}{90}	&	\textcolor{gray}{-}	&	\textcolor{red}{99}	&	\textcolor{gray}{-}	\\
    \hline
    \addlinespace
    \addlinespace
    \multicolumn{7}{c}{LunaTerra addresses on May 8} \\
    \hline
    Address/Motif Center & $C_1$ & $C_4$ & $C_{5a}$ & $C_{5b}$ & $C_6$ & $C_{11}$ \\
    \hline
{\sf \textcolor{gray}{Celsius}}		&	\textcolor{gray}{-}	&	\textcolor{gray}{81}	&	\textcolor{gray}{79}	&	\textcolor{gray}{-}	&	\textcolor{gray}{-}	&	\textcolor{gray}{-}	\\
{\textsf hs0327.eth}		&	\textcolor{gray}{88}	&	\textcolor{gray}{67}	&	\textcolor{gray}{70}	&	\textcolor{red}{96}	&	\textcolor{gray}{82}	&	\textcolor{gray}{-}	\\
{\sf Smart LP: 0x413}		&	\textcolor{gray}{-}	&	\textcolor{gray}{68}	&	\textcolor{gray}{-}	&	\textcolor{gray}{-}	&	\textcolor{red}{95}	&	\textcolor{gray}{-}	\\
{\sf Token Millionaire 1}		&	\textcolor{gray}{85}	&	\textcolor{red}{90}	&	\textcolor{gray}{86}	&	\textcolor{gray}{-}	&	\textcolor{gray}{74}	&	\textcolor{gray}{89}	\\
{\sf Token Millionaire 2}		&	\textcolor{gray}{70}	&	\textcolor{red}{100}	&	\textcolor{red}{99}	&	\textcolor{gray}{-}	&	\textcolor{red}{99}	&	\textcolor{gray}{38}	\\
{\sf masknft.eth}		&	\textcolor{red}{91}	&	\textcolor{red}{91}	&	\textcolor{gray}{82}	&	\textcolor{gray}{-}	&	\textcolor{red}{93}	&	\textcolor{red}{92}	\\
{\sf Heavy Dex Trader}		&	\textcolor{gray}{71}	&	\textcolor{red}{96}	&	\textcolor{gray}{-}	&	\textcolor{gray}{-}	&	\textcolor{gray}{81}	&	\textcolor{gray}{-}	\\
{\sf Oapital}		&	\textcolor{red}{92}	&	\textcolor{gray}{79}	&	\textcolor{gray}{58}	&	\textcolor{gray}{61}	&	\textcolor{gray}{72}	&	\textcolor{red}{93}	\\
{\sf Hodlnaut}		&	\textcolor{gray}{40}	&	\textcolor{red}{99}	&	\textcolor{red}{91}	&	\textcolor{gray}{-}	&	\textcolor{red}{99}	&	\textcolor{gray}{-}	\\
    \hline
    \addlinespace
    \addlinespace
    \multicolumn{7}{c}{LunaTerra addresses on May 9} \\
    \hline
    Address/Motif Center & $C_1$ & $C_4$ & $C_{5a}$ & $C_{5b}$ & $C_6$ & $C_{11}$ \\
    \hline
{\sf \textcolor{gray}{Celsius}}		&	\textcolor{gray}{-}	&	\textcolor{gray}{80}	&	\textcolor{gray}{77}	&	\textcolor{gray}{-}	&	\textcolor{gray}{-}	&	\textcolor{gray}{-}	\\
{\sf hs0327.eth}		&	\textcolor{red}{95}	&	\textcolor{gray}{66}	&	\textcolor{gray}{68}	&	\textcolor{red}{95}	&	\textcolor{gray}{79}	&	\textcolor{gray}{-}	\\
{\sf Smart LP: 0x413}		&	\textcolor{gray}{-}	&	\textcolor{gray}{67}	&	\textcolor{gray}{-}	&	\textcolor{gray}{-}	&	\textcolor{red}{95}	&	\textcolor{gray}{-}	\\
{\sf \textcolor{gray}{Token Millionaire 1}}		&	\textcolor{gray}{83}	&	\textcolor{gray}{89}	&	\textcolor{gray}{85}	&	\textcolor{gray}{-}	&	\textcolor{gray}{73}	&	\textcolor{gray}{88}	\\
{\sf Token Millionaire 2}		&	\textcolor{gray}{67}	&	\textcolor{red}{100}	&	\textcolor{red}{99}	&	\textcolor{gray}{-}	&	\textcolor{red}{99}	&	\textcolor{gray}{88}	\\
{\sf masknft.eth}		&	\textcolor{red}{90}	&	\textcolor{red}{90}	&	\textcolor{gray}{81}	&	\textcolor{gray}{-}	&	\textcolor{red}{92}	&	\textcolor{red}{92}	\\
{\sf Heavy Dex Trader}		&	\textcolor{gray}{70}	&	\textcolor{red}{93}	&	\textcolor{gray}{-}	&	\textcolor{gray}{-}	&	\textcolor{gray}{80}	&	\textcolor{gray}{-}	\\
{\sf Oapital}		&	\textcolor{red}{94}	&	\textcolor{gray}{78}	&	\textcolor{gray}{57}	&	\textcolor{gray}{63}	&	\textcolor{gray}{71}	&	\textcolor{red}{94}	\\
{\sf Hodlnaut}		&	\textcolor{gray}{39}	&	\textcolor{red}{99}	&	\textcolor{red}{90}	&	\textcolor{gray}{-}	&	\textcolor{red}{99}	&	\textcolor{gray}{-}	\\
    \hline
  \end{tabular}
\end{table}

% \smallskip
% \noindent\textbf{K-Core vs. InnerCore.} Compared to traditional graph-$k$-core decomposition where the target is the highest $k$-core, {\textsf InnerCore} with a fixed core target of $\epsilon$ = 0.1 better accentuates the LunaTerra collapse when applied to temporal networks and comparing the sizes of the target cores. 
% From Figure~\ref{fig:lunaTerraCoreSize}, it is evident that the size of the highest graph-$k$-core fluctuates randomly across each day and does not respond to changes in the health of the network. 

% This can be attributed to the fact that graph-$k$-core decomposition only considers the degree of nodes irrespective of the edge weights.  Conversely, {\textsf InnerCore} considers all the provided node features (\S3 and Table~\ref{tab:node_property_functions}) into consideration when computing each node's depth, which includes edge weights in addition to node degree. Therefore, if a node is incident to a single edge of extremely high edge weight, the highest graph-$k$-core will exclude this node; whereas {\textsf InnerCore} will, depending on the target $\epsilon$, maintain it in its {\textsf InnerCore} as the high edge weight offsets the low node degree.

\smallskip
\noindent\textbf{Identify Key Addresses.} Before the LunaTerra collapse, it is reasonable to assume that traders responsible for the collapse would prepare for the anticipated negative consequences by exiting the UST network and entering another reliable stablecoin. In order to capture these transactions of traders converting between different stablecoins, we have included four stablecoins in our network along with UST.  We focus on the unknown addresses that occurred most frequently as motif centers in {\textsf InnerCore}s (defined in \S 3.4) on days immediately before the LunaTerra collapse since they could have influenced the initial phase of the crash.

Generally, a large amount of tokens transferred from one address to another is easily detectable due to the sheer volume.  However, if a trader tries to confiscate detection, the trader could produce multiple transactions with smaller volumes.  Additionally, often in a transaction where one token is exchanged for another, a series of multiple transfers can arise for a single conversion transaction due to interactions with exchanges.\footnote{{\tiny \url{https://etherscan.io/tx/0xa3663b813b2c13a88daeeb5b48b32b7024fc07cbf250f2c2a9318ec1950c9da9}}}
Therefore, a trader is more likely to exhibit both selling and buying behaviors, making the trader a prime candidate as a 3-node motif center.

\smallskip
\noindent\textbf{Ground Truth}. Nansen (\url{https://www.nansen.ai/}) is a prominent blockchain analytics platform that frequently publishes comprehensive analyses of blockchain events, which are followed with great interest by the industry. Nansen.ai conducted a thorough analysis of the LunaTerra collapse in May 2022 and identified 11 important addresses that played central roles in the collapse~\citep{NansenLunaTerra}. We %aim to 
compare the addresses of interest detected by our {\textsf InnerCore} analysis using the centered-motif approach with those identified by Nansen.ai (Table~\ref{tab:nansenAddresses}) as the primary candidates for triggering the collapse.

Exchanges are an intermediary hub to facilitate transfers between traders.  The addresses of exchanges are well-known for this reason, making them not very interesting in our context.  In contrast, addresses that are not exchanges are mostly owned by traders and thus, the existence of such addresses and their edges in a network is a direct consequence of a trader’s activity in the network.  From Table \ref{tab:exchanges}, we observe that motif centers identified from {\textsf InnerCore}s have a high ratio of non-exchange addresses to exchange addresses ($\approx$99\%).  This shows the effectiveness of our method to identify potentially meaningful addresses in a network different from high-traffic exchange addresses.  

In particular, we capture 9 of 11 externally owned addresses ({\textsf EoA}s) in Table \ref{tab:nansenAddresses} identified by Nansen.ai that occurred as center addresses for our motif types (Figure \ref{fig:motifs}) on days immediately leading up to the LunaTerra collapse.  We notice that the {\textsf NF-IAF} score percentile ranks of these addresses are higher compared to that of other center addresses for the same motif type on the same day, indicating that these addresses were important traders contributing to the buy or sell behavior associated with the motif on the day. We surmise the possibility that certain {\textsf EoA}s found by our {\textsf InnerCore} method, coupled with centered-motif analysis, could have been responsible for the initial phase of the collapse.

Recall that in Figure \ref{fig:motifs}, we defined motif centers $C_1$, $C_{5a}$, and $C_{11}$ as exhibiting sell behavior; while motif centers $C_4$, $C_{5b}$, and $C_6$ as exhibiting buy behavior.  It is evident from Table \ref{tab:nansenAddresses} that every motif center on May 8, 2022, has at least one corresponding trader with an {\textsf NF-IAF} score percentile rank above 90.  This suggests that addresses with greater {\textsf NF-IAF} percentiles exhibit a higher buy or sell behavior associated with the particular motif type on the day of the collapse.  
{Specifically, we identify two traders, {\textsf hs0327.eth} and {\textsf Heavy Dex Trader}, as the most likely candidates for influencing the initial phase of the crash, since they had the greatest {\textsf NF-IAF} score percentile increases from May 7 to May 8, 2022 consistently across all their participating motif center types in comparison to other addresses. In addition, we identify the two traders, {\textsf masknft.eth} and {\textsf Oapital}, as key participants throughout the crash, since they are the two addresses with greater {\textsf NF-IAF} percentiles (above 90) occurring consistently across at least two motif types exhibiting sell behavior on days before, during, and after the crash. We identify {\textsf Celsius} as being the least likely trader to have directly impacted the collapse as it is the only address which had score percentiles $<$ 90 across all three days.}

%\smallskip
\noindent\textbf{K-Core vs. InnerCore}. 
% \textcolor{red}{We compare the  addresses identified by {\textsf InnerCore} + centered-motif analysis with those in the highest graph-$k$-core on the May 8 (i.e., one day before the crash) stablecoin temporal network.} 
We notice that graph-$k$-core cannot find any of the 11  addresses indicated by Nansen.ai as prime candidates for triggering the initial phase of %the 
LunaTerra collapse.  In comparison, {\textsf InnerCore} + centered-motif analysis captures potentially anomalous buy and sell behaviors
by identifying 9 of the 11 addresses.   
\vspace{-2mm}
\subsubsection{Experiment 2: Ethereum's Switch to PoS}
Ethereum's transition from Proof-of-Work (PoW) to Proof-of-Stake (PoS) came with many benefits including enhanced security for users and lower energy consumption.  Together, these positives incentivized new traders to participate in the Ethereum network due to increased trust in the blockchain and lower barriers to entry.  The transition occurred in two phases; the first phase was a preparatory hard forking of the blockchain into a PoS structure and the second phase was a finalization of the upgrade.  

\begin{figure}
  \centering
\includegraphics[width=0.45\textwidth]{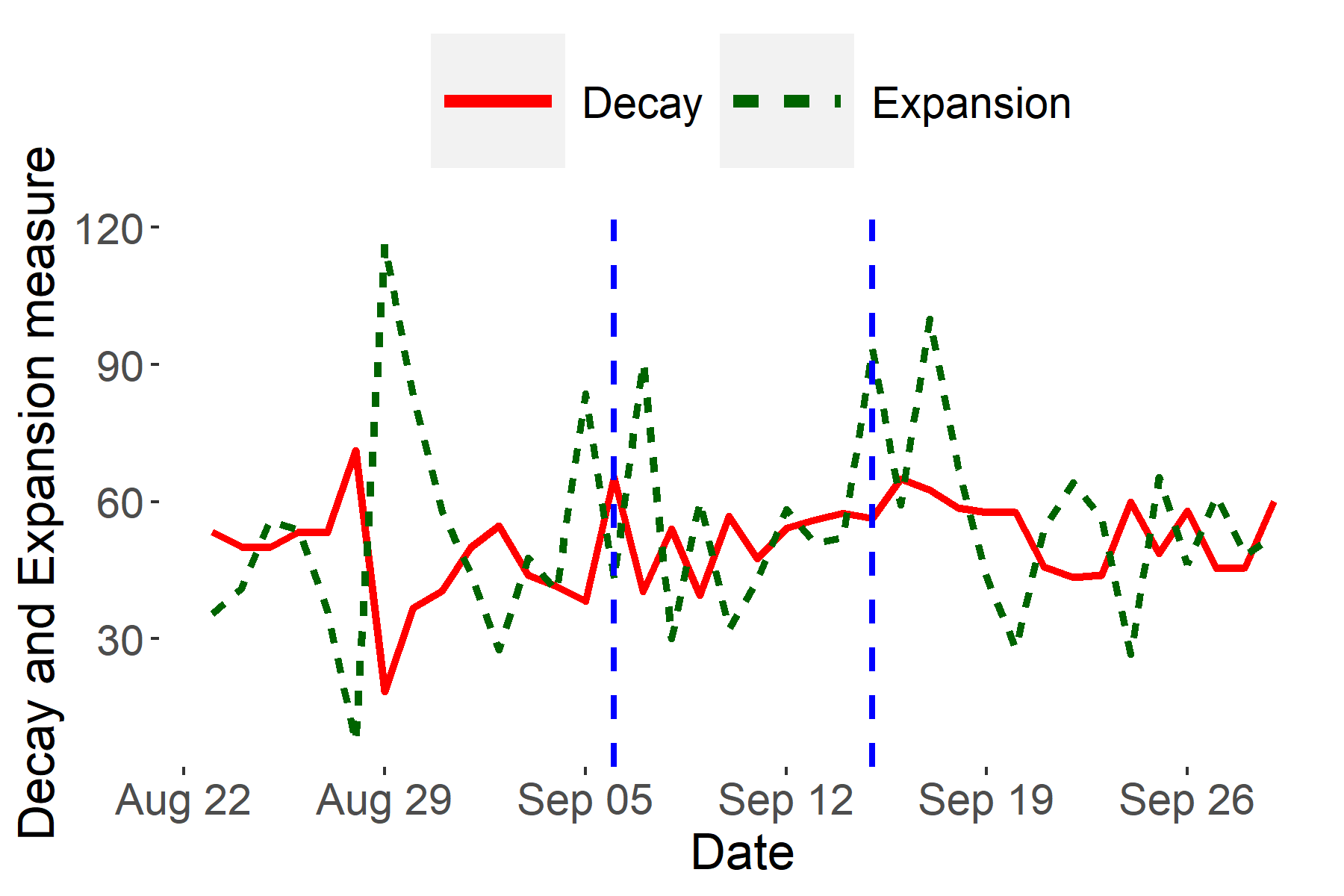}
  \vspace{-3mm}
  \caption{\small Ethereum decay and expansion measures. The move of Ethereum to Proof-of-Stake mining took place in two stages, indicated by 2 vertical blue lines (Sep 6 and 15, 2022). {{An expansion peak on Sep 5, 2022 detects the anomaly one day before the first stage commenced.}} }
  \label{fig:ethDecayExpansion}
  % \vspace{-5mm}
\end{figure}

A pattern of {\em hope} was expected as the upgrade was highly anticipated due to the positives, transparency, and consistent updates regarding the official dates of the upgrade.  
From Figure~\ref{fig:ethDecayExpansion}, we indeed verify this behavioral pattern of {\em hope} characterized by inflated expansion values, coupled with relatively stable decay values, on three separate occasions.  The first occurrence of {\em hope} is observed approximately a week before the first phase of the upgrade took place.  It was around this time, the end of August 2022, that official news regarding the concrete dates of when the upgrade would be expected to take place was released to the public.  We observe a surge of new hopeful traders participating in the Ethereum network and a significant dip in existing traders leaving the network in  
 anticipation of the upgrade.  The other two instances of {\em hope} are seen during the immediate days surrounding and between each of the phases of the upgrade.  These occurrences provide insight into the market sentiment during the upgrade as positive and the overall transition of Ethereum to PoS as being well-received by traders.

\begin{figure}
  \centering
\includegraphics[width=0.45\textwidth]{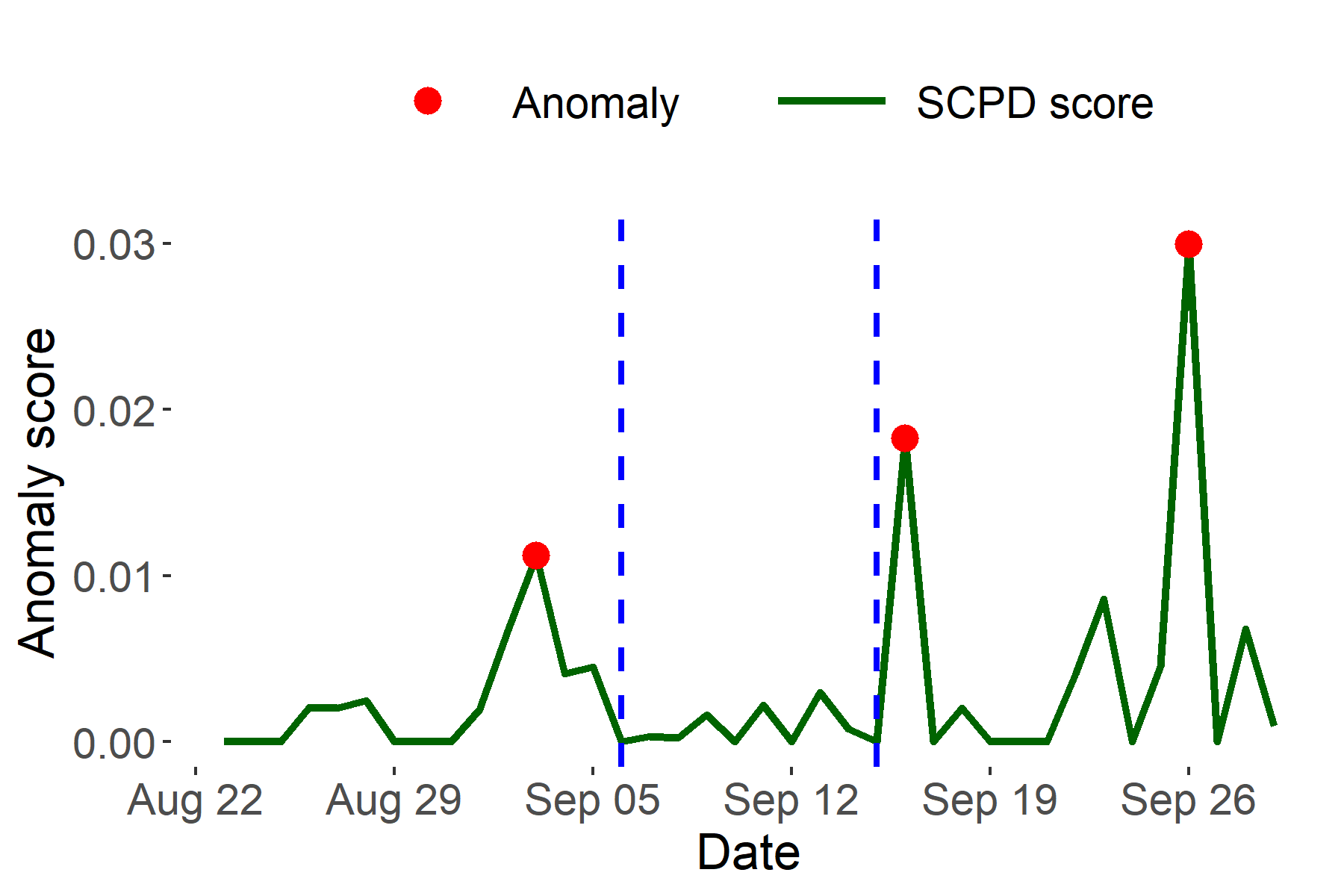}
  \vspace{-3mm}
  \caption{\small Ethereum anomalous days identified by {\textsf SCPD}. Compared to decay and expansion measures by {\textsf InnerCore}, {\textsf SCPD} less accurately captures the two phases of Ethereum's transition to POS occurring on Sep 6 and 15, 2022.}
  \label{fig:ethSCPD}
  % \vspace{-5mm}
\end{figure}

\smallskip
\noindent\textbf{SCPD vs. InnerCore}. 
We next apply {\textsf SCPD} on the Ethereum transaction network to compare against our expansion and decay results.  From Figure~\ref{fig:ethSCPD}, we notice that {\textsf SCPD} less accurately captures the two phases of Ethereum's transition to POS occurring on Sep 6 and 15, 2022.  {\textsf SCPD} identifies Sep 9 and 16 as anomalous, which are two days before the first phase and one day after the second phase, respectively, of Ethereum's transition to POS.  In contrast, our expansion measures in Figure~\ref{fig:ethDecayExpansion} more accurately capture the phases of Ethereum transition to POS by producing a peak on Sep 4, one day before the first phase, and on Sep 15, the same day of the second phase. {It is evident {\textsf InnerCore} detects the second phase of the switch on the day of the event, whereas {\textsf SCPD} can only detect the event after it has occurred.}   Therefore, {\textsf InnerCore} expansion measures more accurately detect an anomaly on days when a significant event actually unfolded.

\subsubsection{Experiment 3: USDC's Temporary Peg Loss}
\label{sec:exp3}
On May 11th, 2023, a significant event unfolded in the stablecoin market as Circle's stablecoin, USDC, experienced a temporary loss of its peg, plummeting to a concerning value of 87 cents.\footnote{{\tiny \url{https://coinmarketcap.com/currencies/usd-coin/}}} The abrupt collapse of Silicon Valley Bank, which held over 3 billion of Circle's reserves, triggered panic among traders. Fearing a collapse, many traders liquidated their USDC holdings and sought refuge in alternative stablecoins like MakerDAO's DAI. 
% \textcolor{red}{Notably, USDC differs from Terra's UST, which had algorithmic pegging to Luna and had suffered a complete collapse approximately a year earlier. USDC, in contrast, maintains a strong collateralization to fiat reserves, marking a crucial distinction between the two stablecoins.} 

\begin{figure}
  \centering      
  \includegraphics[width=0.45\textwidth]{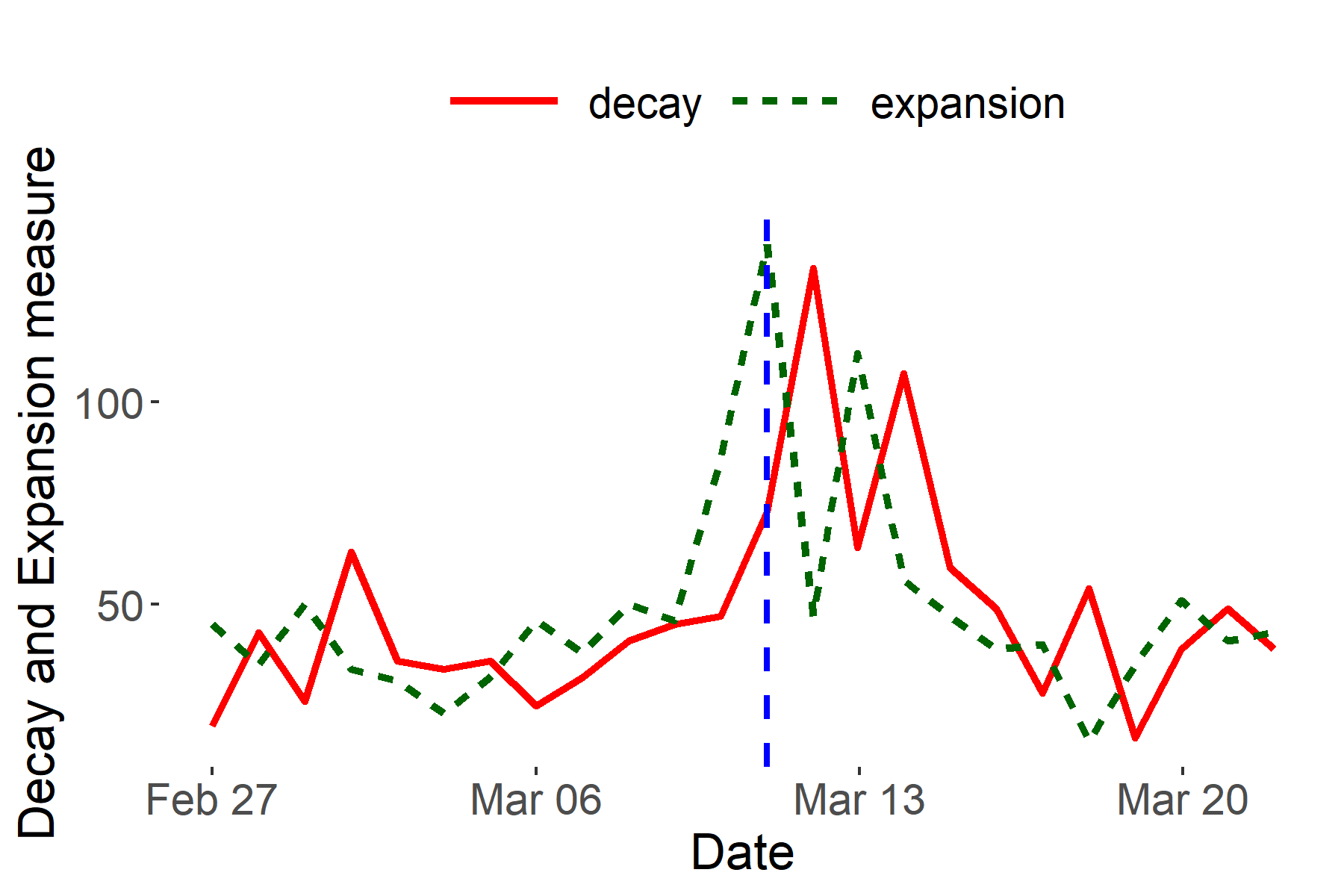}
  \vspace{-3mm}
  \caption{\small USDC decay and expansion measures. On Mar 11, 2023 (shown with the vertical blue line), USDC loses its \$1 peg and falls to as low as 87 cents. {{An expansion peak detects the anomaly on the day the event transpires.}}}
  \label{fig:usdcExpansionDecay}
  % \vspace{-5mm}
\end{figure}

By analyzing the expansion and decay measures surrounding the incident, we realize how traders responded differently to this event. Figure~\ref{fig:usdcExpansionDecay} shows a sudden surge in expansion on May 11th, 2023, attributing to a wave of traders liquidating their USDC holdings in response to the stablecoin's all-time low value of 87 cents. In the subsequent three days following the temporary loss of USDC's peg, a distinct series of behavioral patterns emerged, characterized by alternating signals of {\em despair}, {\em hope}, and {\em despair} again, before eventually stabilizing. During this three-day period, Circle's reassurances regarding the recovery of lost reserves gradually restored trust among its traders. This is evident through the decreasing extent of {\em despair} patterns observed on the 12th and 14th.

In summary, traders' reactions were initially marked by panic and a rush to sell USDC, causing a surge in expansion. However, as Circle provided updates on their efforts to recover the lost reserves, a sense of hope permeated the market, leading to a decline in the extent of despair patterns. Ultimately, the stablecoin regained stability, with expansion and decay returning to typical levels.

\begin{figure}
  \centering      
  \includegraphics[width=0.45\textwidth]{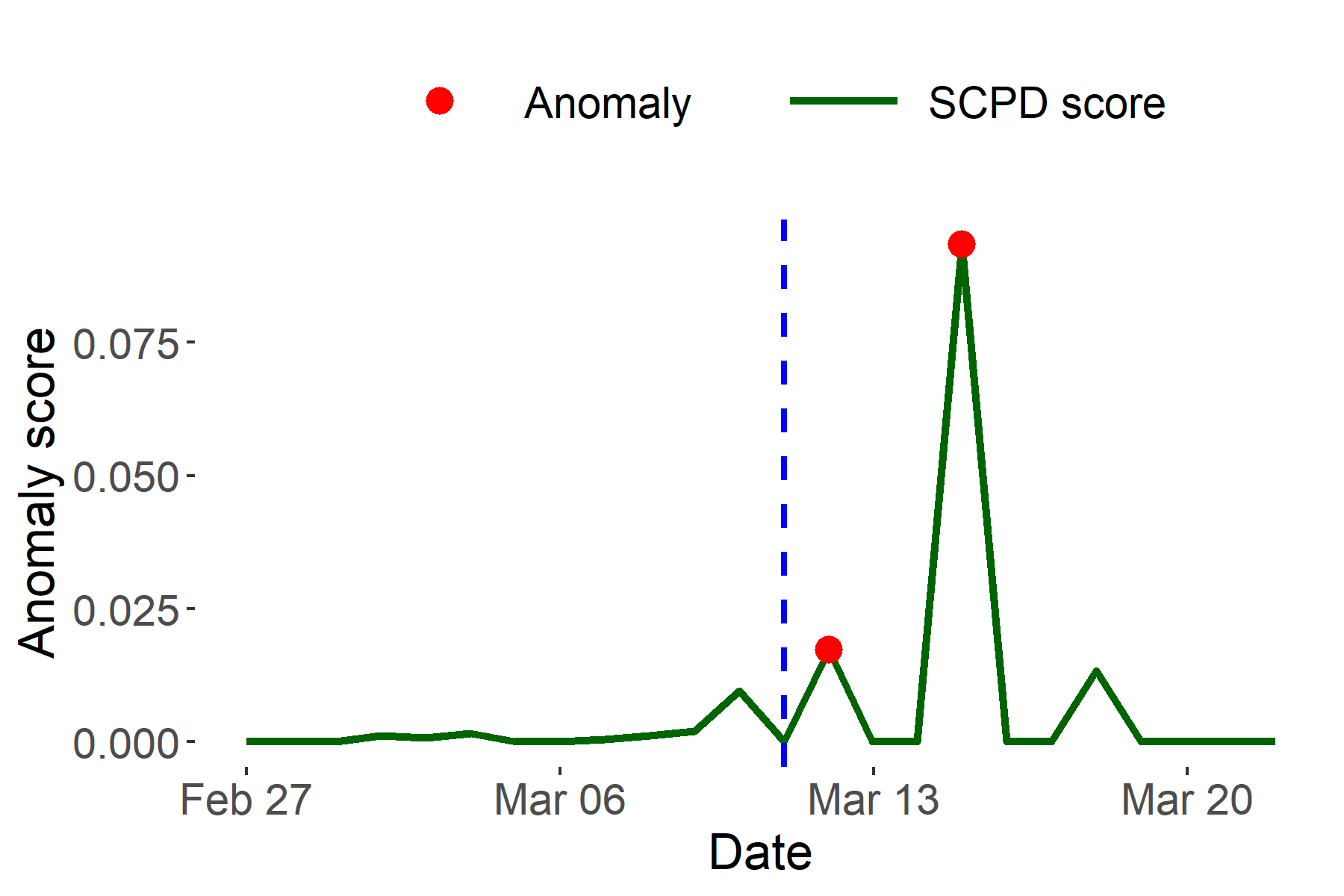}
  \vspace{-3mm}
  \caption{\small USDC anomalous days identified by {\textsf SCPD} . Compared to decay and expansion measures by {\textsf InnerCore}, {\textsf SCPD}  less accurately captures USDC's temporary peg loss occurring on Mar 11, 2023.}
  \label{fig:usdcSCPD}
  % \vspace{-5mm}
\end{figure}

\smallskip
\noindent\textbf{SCPD vs. InnerCore}. 
% \spara{Effectiveness Experiment 3: SCPD to the USDC network.}
We also apply {\textsf SCPD}  to the USDC network in order to compare with our decay and expansion results.  From Figure~\ref{fig:usdcSCPD}, we observe that {\textsf SCPD}  less accurately captures USDC's temporary peg loss occurring on Mar 11. {\textsf SCPD}  identifies Mar 12 and 15 as anomalous which are one day and four days, respectively, after USDC's peg loss.  Conversely, our expansion measures in Figure~\ref{fig:usdcExpansionDecay} accurately capture USDC's peg loss by producing a prominent peak on Mar 11. {It is evident that {\textsf InnerCore} detects the temporary peg loss on the day of the event, whereas SCPD can only detect the event after it has occurred.}
Clearly, our {\textsf InnerCore} expansion measures more accurately indicate an anomaly on days when a significant event occurred.
 

%% file: sections/07_conclusion.tex
%\vspace{-2mm}
\section{Conclusions}
\label{sec:conc}
%In this article, 
We have introduced {\textsf InnerCore}, which utilizes data depth-based core discovery to identify the influential nodes in temporal blockchain token networks. Furthermore, we have proposed two metrics, {\textsf InnerCore} expansion and decay, that provide a sentiment indicator for the networks and explain trader mood. 

Finally, with a centered-motif analysis in the {\textsf InnerCore}, we detected market manipulators and e-crime behavior. 
The scalability and computational efficiency of {\textsf InnerCore} discovery make it well-suited for analyzing large temporal graphs, including those found in Ethereum transaction and stablecoin networks. Our experiments, which compare our findings against external ground truth, baselines, and state-of-the-art attributed change detection
approach in dynamic graphs, show that {\textsf InnerCore} efficiently extracts useful information from large networks compared to existing methods. 
In future, we shall use {\textsf InnerCore} to explore network robustness against Decentralized Finance (DeFi) attacks.